\PassOptionsToPackage{dvipsnames,svgnames,table,x11names}{xcolor}
\documentclass[12pt, letterpaper]{article}
\usepackage{pdfpages}

\usepackage{ragged2e}
\usepackage{url} 
\usepackage{graphicx} 
\usepackage{caption} 
\usepackage[inkscapelatex=false]{svg}
\usepackage{blindtext}
\usepackage{geometry}
\usepackage{listings}
\usepackage{verbatim}
\usepackage[english]{babel}
\usepackage[utf8x]{inputenc}
\usepackage{tikz,lmodern}
\usepackage[most]{tcolorbox}
\usepackage{listings}
\usepackage{subcaption} 
\usepackage{float}              
\floatstyle{plaintop}           
\restylefloat{table}
\usepackage{booktabs}
\usepackage{longtable}
\usepackage{adjustbox}
\usepackage{graphicx}
\usepackage{pdflscape} 
\usepackage{geometry}  

\usepackage{booktabs}
\usepackage{dcolumn}

\lstdefinelanguage{json}{
    basicstyle=\ttfamily\footnotesize,
    numbers=left,
    numberstyle=\scriptsize,
    stepnumber=1,
    numbersep=8pt,
    showstringspaces=false,
    breaklines=true,
    frame=lines,
    backgroundcolor=\color{gray!10},
    literate=
     *{0}{{{\color{blue}0}}}{1}
      {1}{{{\color{blue}1}}}{1}
      {2}{{{\color{blue}2}}}{1}
      {3}{{{\color{blue}3}}}{1}
      {4}{{{\color{blue}4}}}{1}
      {5}{{{\color{blue}5}}}{1}
      {6}{{{\color{blue}6}}}{1}
      {7}{{{\color{blue}7}}}{1}
      {8}{{{\color{blue}8}}}{1}
      {9}{{{\color{blue}9}}}{1}
}

\usepackage{tabularx}
\usepackage{caption}
\usepackage{titlesec}

\usepackage{booktabs}

\usepackage{siunitx}                        
\usepackage{threeparttable}               

\usepackage{makecell}
\usepackage{adjustbox}
\usepackage{helvet}
\usepackage{times}
\usepackage{placeins}
\usepackage{multirow}
\usepackage{subfig}
\usepackage{subcaption}
\usepackage{float}
\usepackage{multicol}
\usepackage{supertabular}
\usepackage{booktabs}
\usepackage[framemethod=TikZ]{mdframed}
\usepackage{fullpage}
\usepackage[switch]{lineno}
\usepackage{amssymb}
\usepackage{amsmath}
\usepackage{rotating}
\usepackage{array}
\usepackage{mathtools}
\usepackage[ruled]{algorithm2e}
\usepackage{algorithmic}
\usepackage{bm}
\usepackage{comment}
\usepackage{enumitem}
\usepackage{graphics}
\usepackage{latexsym}
\usepackage{mathrsfs}
\usepackage{morefloats}
\usepackage{lineno}
\usepackage{nicefrac}
\usepackage{authblk}
\usepackage[english]{babel}
\usepackage{blindtext}
\usepackage{siunitx}
\usepackage{xr}

\usepackage{dcolumn}
\usepackage{todonotes}

\usepackage{hyperref}
\usepackage{xr-hyper}
\usepackage{cleveref}
\makeatletter
\providecommand{\sf@counterlist}{} 
\makeatother

\usepackage{longtable}
\usepackage{array}
\usepackage{appendix}
\usepackage{setspace} 
\usepackage{times}
\usepackage{enumitem}
\usepackage{amssymb} 
\usepackage{tikz}    
\usepackage{hyperref}

\usepackage{dsfont}
\usepackage{array}

\usepackage{comment}
\usepackage{ragged2e}
\usepackage{placeins}
\usepackage{tabularx}
\usepackage{multirow}
\usepackage{rotating}
\usepackage{mathtools}
\usepackage{arydshln}
\usepackage{longtable}
\usepackage{float}
\usepackage{multicol}
\usepackage{amssymb}
\usepackage{url}
\usepackage{svg}
\newcommand{\toneA}{\textcolor{green!50!black}{\textbf{Appreciative}}}
\newcommand{\toneC}{\textcolor{blue!70!black}{\textbf{Constructive-Analytical}}}
\newcommand{\toneQ}{\textcolor{orange!80!black}{\textbf{Questioning}}}
\newcommand{\toneE}{\textcolor{red!70!black}{\textbf{Critical-Evaluative}}}

\newcommand{\radiobtn}{\tikz\draw[fill=white] (0,0) circle [radius=0.12cm];}

\usepackage{amsmath}
\usepackage{supertabular}
\usepackage{titlesec}
\usepackage{makecell}
\usepackage{adjustbox}
\usepackage{graphicx}
\usepackage{dcolumn}
\usepackage{lineno}
\usepackage{hyperref}
\usepackage{cleveref}
\usepackage{xr}   
\graphicspath{{supplementary_figures/}{../supplementary_figures/}}

\makeatletter
\newcommand*{\addFileDependency}[1]{
  \typeout{(#1)}
  \@addtofilelist{#1}
  \IfFileExists{#1}{}{\typeout{No file #1.}}
}
\makeatother

\newcommand*{\myexternaldocument}[1]{
    \externaldocument{#1}
    \addFileDependency{#1.tex}
    \addFileDependency{#1.aux}
}
\usepackage{subfiles} 
\myexternaldocument{supplementary}

\graphicspath{{main_figures/}{../main_figures/}}

\title{\textbf{Disparities in Peer Review Tone and the Role of Reviewer Anonymity}}
\author[1]{Maria Sahakyan}
\author[1*]{Bedoor AlShebli}

\affil[1]{\normalsize 
New York University Abu Dhabi, Social Science Division}

\affil[*]{\small corresponding author email:\ bedoor@nyu.edu}
\date{}

\begin{document}
\maketitle
\vspace{-2em}
\begin{abstract}
\noindent The peer review process is often regarded as the gatekeeper of scientific integrity, yet increasing evidence suggests that it is not immune to bias. Although structural inequities in peer review have been widely debated, much less attention has been paid to the subtle ways in which language itself may reinforce disparities. This study undertakes one of the most comprehensive linguistic analyses of peer review to date, examining more than 80,000 reviews in two major journals. Using natural language processing and large-scale statistical modeling, it uncovers how review tone, sentiment, and supportive language vary across author demographics, including gender, race, and institutional affiliation. Using a data set that includes both anonymous and signed reviews, this research also reveals how the disclosure of reviewer identity shapes the language of evaluation. The findings not only expose hidden biases in peer feedback, but also challenge conventional assumptions about anonymity’s role in fairness. As academic publishing grapples with reform, these insights raise critical questions about how review policies shape career trajectories and scientific progress.
\end{abstract}

\section*{Introduction}

Peer review remains a cornerstone of scholarly publishing, essential for safeguarding the quality, credibility, and integrity of scientific research. Despite its fundamental role, the peer review process is still poorly understood and continues to provoke debate regarding its purpose, effectiveness, and fairness~\cite{tennant2020limitations}. Growing evidence suggests that peer review is susceptible to social biases that may undermine objectivity and equity in the evaluation of manuscripts~\cite{lee2013bias}. Moreover, recent work highlights systemic shortcomings, including low inter-reviewer agreement, procedural inefficiencies, and limited transparency, which further challenge the integrity of the process~\cite{aczel2025present}. As science becomes increasingly global and interdisciplinary, there is an urgent need to clarify the normative goals of peer review, evaluate alternative models, and develop empirically grounded reforms to mitigate bias and improve the consistency and fairness of scientific evaluation.

At its core, peer review is intended to enhance the quality of scientific research by identifying methodological flaws, offering constructive feedback, and flagging potentially misleading claims. However, it has faced persistent criticism for its inefficiencies, lack of transparency, and vulnerability to bias~\cite{aczel2025present, schwartz2009peer, bohannon2013s, tomkins2017reviewer, smith2006peer}. Despite these concerns, the process continues to receive broad support from researchers and journal stakeholders~\cite{mulligan2013peer, jefferson2006editorial}. Nevertheless, concerns about unprofessional reviewer behavior remain, with survey-based studies documenting instances of inappropriate or biased comments related to gender, race, ethnicity, or nationality of authors: issues that can be particularly damaging for early-career researchers~\cite{silbiger2019unprofessional, beaumont2019peer, hyland2020work}.

To begin unpacking the specific dimensions of bias, we first consider gender. Research on gender bias in peer evaluation reveals a complex and often contradictory landscape. Some studies suggest that female first authors receive less polite or more critical feedback compared to their male counterparts, while papers led by female senior authors may be evaluated more favorably~\cite{verharen2023chatgpt}. Additional work points to the underrepresentation of women in high-prestige authorship and reviewer roles, along with evidence that women may be subject to harsher critique~\cite{budden2008double}. In contrast, large-scale linguistic analyses using lexical tools such as LIWC have found that reviewer gender and review model (e.g., single- vs. double-anonymous)  have only a limited effect on the linguistic style of peer reviews~\cite{buljan2020large, squazzoni2021peer}. A recent meta-analysis further complicates the picture, reporting gender parity in journal acceptances, grant funding, and hiring outcomes, but persistent disparities in teaching evaluations and salaries~\cite{ceci2023exploring}. 

Beyond gender, geographic and institutional biases represent some of the most well-documented and enduring forms of inequity in peer review. Numerous studies have shown that authors based at institutions in non-Western, non-English speaking, or lower income regions face systematically lower acceptance rates, even when the quality of their submissions remains constant~\cite{smith2023peer, casadevall2014causes}. Reviewers frequently favor submissions from high-income countries, elite institutions, and well-known researchers, especially under single-anonymous review conditions~\cite{tomkins2017reviewer}. Experimental and simulated evaluations confirm that affiliation bias persists even when identical manuscripts are evaluated, with both human reviewers and large language models rated submissions from prestigious institutions more favorably~\cite{von2024affiliation, kowal2022impact}. Although double-anonymous review can mitigate some of these disparities~\cite{cuskley2020double, tomkins2017reviewer, sun2022does}, its adoption remains limited~\cite{smith2023peer}, and it does not eliminate all forms of bias. Reviewers may still respond unfavorably to manuscripts that challenge dominant paradigms or address stigmatized topics, such as structural racism, thereby constraining epistemic diversity even when author identities are concealed~\cite{strauss2023racism}. Large-scale analyses of editorial workflows further reveal that geographic and prestige-related disparities contribute to the disproportionate publication of work from elite institutions and regions, while marginalizing contributions from less-resourced contexts and underrepresented research agendas~\cite{kulal2025unmasking, zhang2022investigating}. 
These findings suggest that reforms aimed at enhancing transparency and accountability in the peer review process, such as increased openness around reviewer identities, clearer documentation of review rationale, and mechanisms for auditing bias, may offer promising pathways toward greater equity and epistemic inclusion.

Building on these documented disparities, the present study introduces a multilevel framework for evaluating peer review content using advanced natural language processing (NLP) techniques. Although prior research has underscored the subjectivity and variability inherent in reviewer assessments, standardized approaches for analyzing the linguistic and structural features of peer review remain limited. To address this gap, we propose a three-tier analytical pipeline that integrates n-gram analysis with large language models (LLMs), including SciBERT and OpenAI’s GPT-4o, to systematically examine reviewer language, evaluative tone, and feedback structure. All reviews in our dataset correspond to the first round of reviews for accepted papers, allowing us to examine potential bias in reviews of ultimately successful submissions while holding outcome constant. In other words, these are the reviews where bias should, theoretically, be the lowest. As such, if bias is evident even in this subset, it likely persists, if not intensifies, in the broader set of rejected submissions. Furthermore, this methodology enables scalable, fine-grained analysis of review texts and provides new insights into the social and rhetorical dynamics that shape scholarly evaluation.

To implement this framework, our analysis began by fine-tuning SciBERT~\cite{beltagy2019scibert}, a domain-specific variant of BERT optimized for scientific text, to conduct sentiment classification on full-length peer review documents. This model effectively captures nuanced sentiment in complex academic language~\cite{bharti2023politepeer, ghosal2022peer}. We then examined contextual linguistic patterns through n-gram analysis, focusing on frequently occurring bigrams and trigrams~\cite{falk2019language, al2022analysis, sun2024textual}. By comparing n-gram distributions across author characteristics, including gender, race, geographic affiliation, and institutional prestige, we identified subtle but meaningful variations in evaluative phrasing. Finally, we defined a set of tone categories informed by prior psychological and computational research~\cite{mehrabian1971silent, picard1997w, steffens2021emotional, ramachandran2017automated, jawaid2006characteristics}, and used OpenAI’s GPT-4o to classify peer review sentences accordingly. Treating tone as an outcome variable in regression models allowed us to systematically assess how reviewer tone varies as a function of author attributes.

Together, these components link sentiment, linguistic patterns, and tone in peer reviews to the demographics of the corresponding authors. By illuminating potential systemic biases, it highlights the need for more equitable and transparent approaches to scholarly evaluation. Our findings offer empirical grounding for policy discussions aimed at strengthening fairness and accountability in peer review.

Finally, we turn to the role of reviewer anonymity, a dimension that has received growing attention in debates around transparency and accountability in peer review. Prior studies suggest that anonymous reviews tend to be more candid and, at times, more critical~\cite{kourilova1996interactive}. Building on this work, we explore how identity disclosure influences the tone and content of peer evaluations. To examine this dynamic, we compared reviews by anonymous and disclosed reviewers, analyzing how the presence or absence of reviewer identity influences the language and style of evaluation. This comparison helps to disentangle the effects of anonymity from other reviewer characteristics, providing clearer insight into how transparency shapes peer review behavior.

\section*{Results}

Although structural biases in academia are well documented, fewer studies have examined how linguistic patterns in peer review contribute to these inequities. Understanding how review tone, sentiment, and expressions of support vary across author demographics is critical for designing fairer evaluation systems. To address this gap, we applied natural language processing techniques to systematically analyze the language of peer reviews.

We conducted a comprehensive analysis of a dataset comprising 32,862 academic manuscripts, published between 2019 and 2024, along with their corresponding peer reviews, sourced from two prominent open-access platforms: Nature Communications and PLOS One. These journals were selected for their multidisciplinary scope and their practice of openly publishing peer reviews alongside manuscripts, a transparency that provided a unique opportunity to investigate biases in signed versus anonymous reviews. Both journals allow reviewers the option to disclose their identities or remain anonymous, enabling a direct exploration of how reviewer anonymity influences feedback. Their reputation for rigorous peer review and commitment to high-quality research further validated their suitability for our study.

Our analysis focused exclusively on the first round of peer review reports, as subsequent rounds were intentionally omitted. Additionally, we excluded reviews of rejected manuscripts since these reviews are not publicly accessible. Including such reviews would also have introduced bias into our analysis, as rejected manuscripts are more likely to receive harsher critiques regardless of reviewer anonymity. By focusing only on the first round of reviews for accepted manuscripts, we ensure a consistent and fair basis for studying potential biases between anonymized and signed reviews.

\begin{figure}[!ht]
    \centering
    \includegraphics[width=0.95\textwidth]{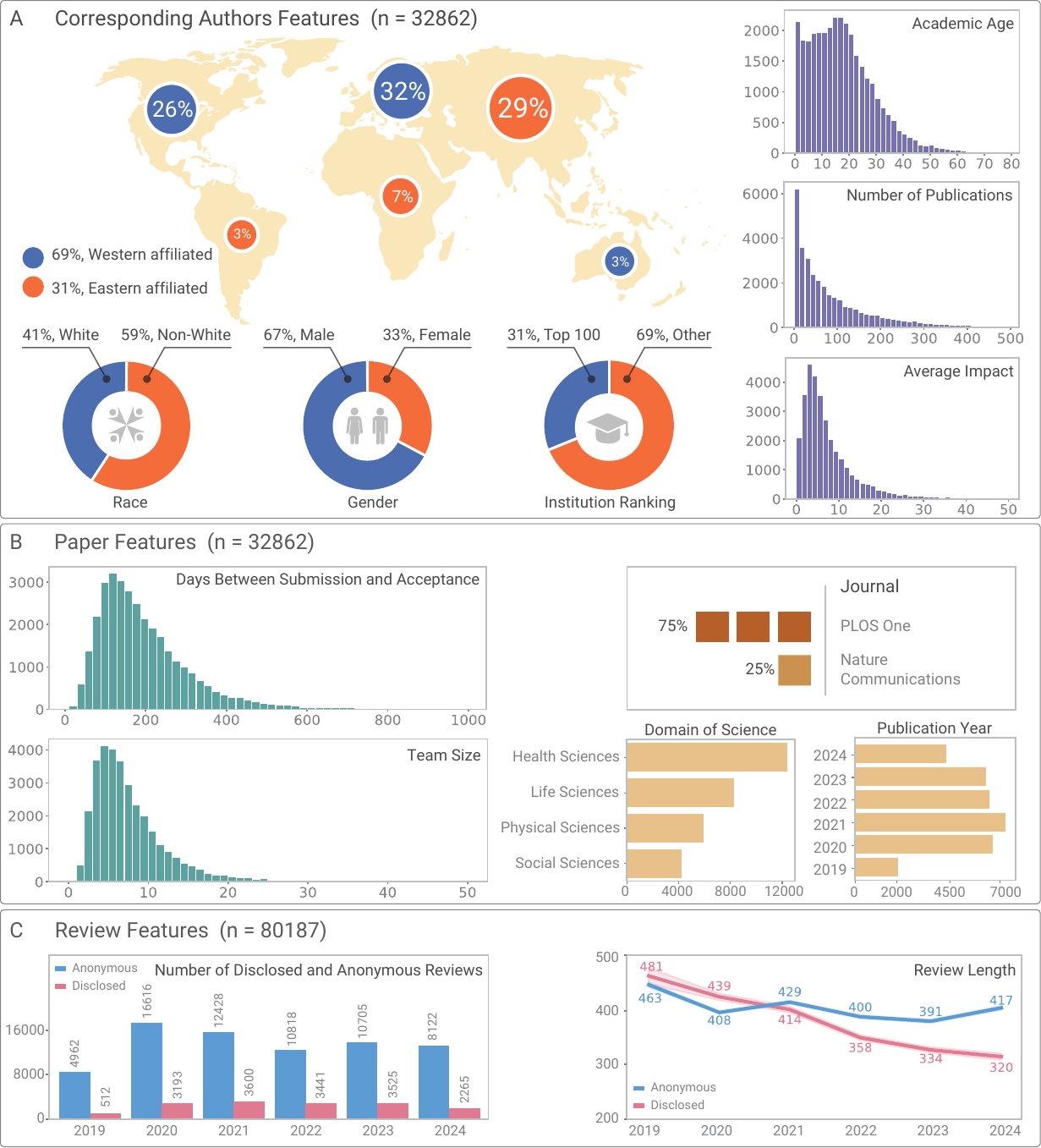}
\caption{\textbf{Overview of the dataset.} The figure presents key characteristics of our dataset, including: \textbf{(a)} the distribution of corresponding author attributes such as institutional affiliation, gender, race, and academic age; \textbf{(b)} paper-specific features like field of science, team size, and submission timelines (for a more detailed breakdown of the fields of science, refer to Supplementary Figure~\ref{fig:suppl_fields_of_science}); and \textbf{(c)} review-related metrics, including anonymity and length. }
    \label{fig:infographics}
\end{figure}

To prepare for our analysis, we extracted and engineered a range of features related to corresponding authors, including gender, race, affiliation rank, country of affiliation, academic age, and more. Additionally, we incorporated paper-specific features such as the field of science, publication year, and review duration, as well as review-related attributes like review length. These features provide critical contextual insights into the characteristics of corresponding authors, the publication itself and the peer review process. A comprehensive list of these features, along with detailed information about our data collection process, is available in the Data section of Materials and Methods. An overview of the distribution of our final dataset, including the number of manuscripts, total reviews, and signed reviews across both journals, is presented in Table~\ref{tab:datadescription}. Meanwhile, Figure~\ref{fig:infographics} provides a detailed description of the distributions of corresponding author attributes, manuscript characteristics, and review features.

To investigate linguistic patterns in peer reviews, we performed a three-part analysis across author demographics, including gender, race, institutional prestige, and geographic location. First, we fine-tuned SciBERT, a transformer model optimized for scientific text, to classify the sentiment of review sentences as positive, neutral, or negative. Second, we performed a trigram analysis to identify frequently occurring linguistic structures that may reflect implicit biases. Third, we used a large language model (LLM) to evaluate each sentence for the presence and strength of four tone types: appreciative, constructive-analytical, questioning, and critical-evaluative. Finally, we used linear regression models to assess how these tones vary based on author characteristics and reviewer anonymity.

\subsection*{Sentiment Differences Across Author Groups}

To understand how reviewers' tone may vary across different types of authors, we begin by examining sentiment in peer review texts. Sentiment analysis provides a coarse but informative lens into the overall valence of reviewer language, whether it is positive, neutral, or negative, offering a starting point for identifying disparities in evaluative tone. We used a SciBERT-based classifier fine-tuned on peer review texts to assign sentiment labels to individual reviews; details of the model and validation procedure are provided in the Materials and Methods section.

\begin{figure}[!h]
    \centering
    \includegraphics[width=1\textwidth]{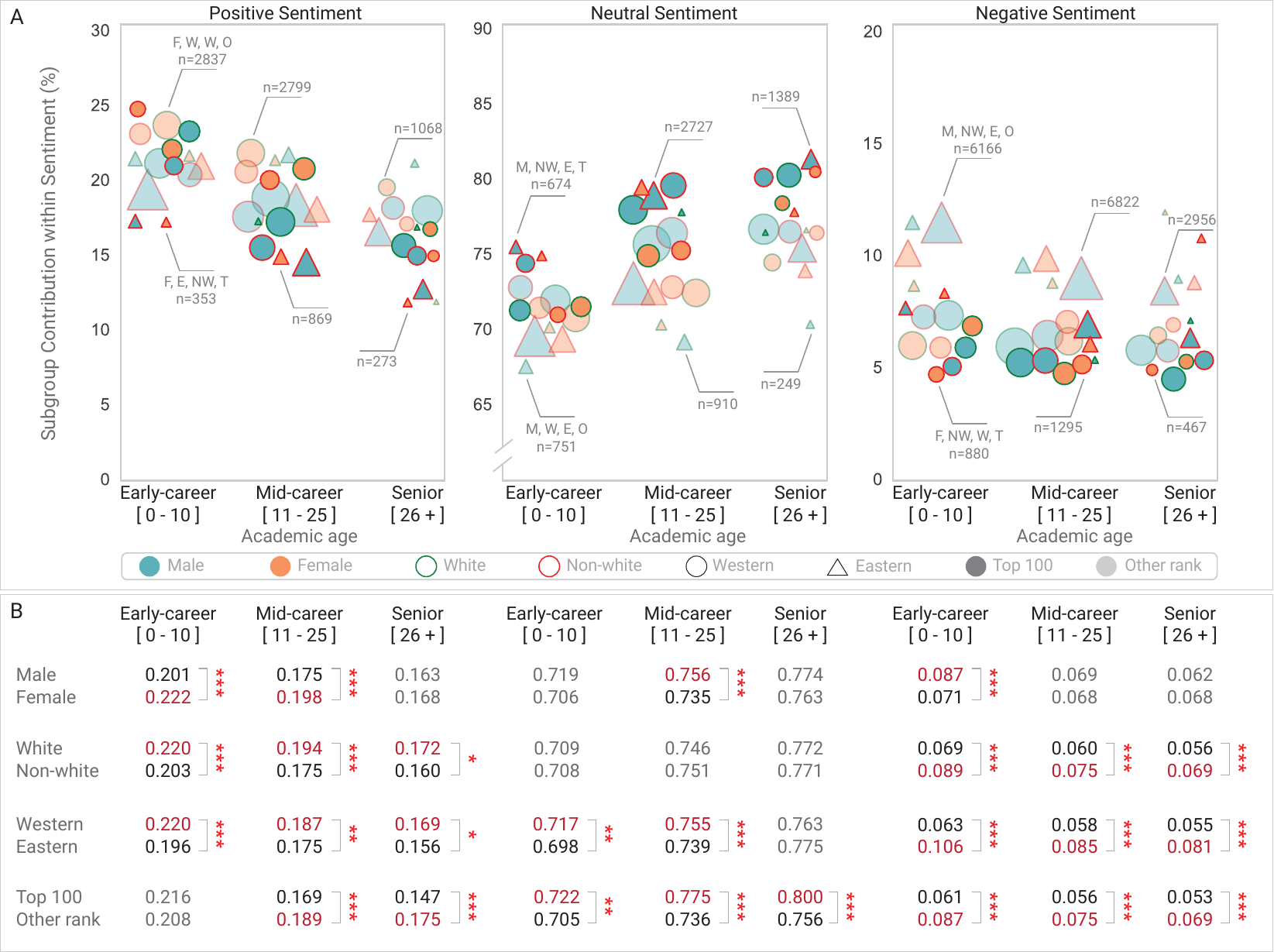}
   \caption{\textbf{Sentiment analysis of peer reviews across author subgroups and academic age bins.}\\
\textbf{(a)} Distribution of subgroup contributions to each sentiment category (Positive, Neutral, Negative) across academic age bins: [0,10], [11,25], and [26+]. Marker color indicates gender (Male vs. Female), border color reflects race (White vs. Non-white), marker shape encodes region of affiliation (Western vs. Eastern), and fill opacity represents institutional rank (Top 100 = opaque, Other rank = semi-transparent). Marker size corresponds to the total number of peer reviews received by papers whose corresponding authors belong to a given subgroup (defined by gender, race, region, and rank) within the respective academic age bin, aggregated across all sentiment categories. The y-axis shows each subgroup’s percentage contribution to that sentiment category. Six representative subgroups are annotated on the plots for illustration. The full subgroup-level review counts used to determine marker sizes are available in the Supplementary Table~\ref{tab:subgroup_sizes_sentiment}. \textbf{(b)} Results of two-sample proportion Z-tests comparing sentiment proportions between subgroup pairs (e.g., Male vs. Female) within each academic age bin. Reported values are subgroup-level proportions for each sentiment. Red numbers indicate the significantly dominant group (i.e., the group with the higher proportion). Asterisks denote significance levels (*** $p < 0.001$, ** $p < 0.01$, * $p < 0.05$). Faded numbers represent comparisons that did not reach statistical significance.}
\label{fig:sentiment_analysis}
\end{figure}

In Figure~\ref{fig:sentiment_analysis}a, we visualize how sentiment is distributed across 16 subgroups of corresponding authors, defined along four dimensions: gender (Male, Female), race (White, Non-white), region of institutional affiliation (Western, Eastern), and institutional rank (Top 100, Other). These are shown across three stages of academic seniority: early-career (0–10 years), mid-career (11–25 years), and senior (26+ years), to account for potential shifts in how authors are evaluated over the course of their careers. Each marker represents a subgroup, with visual encodings indicating group characteristics: color for gender, border color for race, shape for region, and opacity for institutional rank. Marker size corresponds to the number of peer reviews linked to that subgroup, as detailed in Supplementary Table~\ref{tab:subgroup_sizes_sentiment}.

The y-axis reflects the proportion of reviews within each subgroup that fall into a given sentiment category. For example, among early-career authors in the “Female, White, Western, non-Top 100” subgroup (F, W, W, O; n = 2,837), roughly 23.4\% of their reviews are classified as positive, with the remainder split between neutral (approximately 70.7\%) and negative (around 5.9\%) sentiment. Importantly, the values on the y-axis represent within-subgroup sentiment distributions, each point shows how a single group’s reviews are divided across sentiment categories, rather than how sentiment is distributed across all groups within an academic age bin.

We find that across all academic age groups, neutral sentiment was the most common tone. This tendency was particularly pronounced in reviews addressed to senior authors affiliated with top 100 institutions, suggesting a relationship between institutional prestige, academic seniority, and more measured reviewer language. In contrast, authors from non-top 100 institutions experienced a more polarized sentiment distribution, receiving relatively more reviews with both positive and negative tone.

Positive sentiment was more prevalent in reviews of early-career authors, with a decreasing trend across mid- and senior-career stages. Female, white, and western-affiliated authors generally received a higher proportion of positive sentiment, especially in early career stages. Negative sentiment, while infrequent overall, was more commonly directed toward early-career authors who were non-white, eastern-affiliated, or from non-top 100 institutions. Among early-career authors, male authors also received more negative sentiment compared to female counterparts.

To statistically assess these differences, we performed two-sample proportion Z-tests for each sentiment category across gender, race, regional affiliation, and institutional rank within each academic age bin. The results are presented in Figure~\ref{fig:sentiment_analysis}b. Significant differences are marked with asterisks, and the group with the higher sentiment proportion is highlighted in red.

The statistical analysis confirmed that disparities in reviewer sentiment align with demographic and institutional characteristics of authors. Female authors were more likely to receive positive sentiment early in their careers, while male authors were more exposed to negative sentiment in early stages and neutral sentiment in mid-career. Reviews of white and western authors received more positive tones, whereas reviews of non-white and eastern authors were more frequently negative. Prestige also played a role: top 100 affiliated authors received more neutral reviews, perhaps suggesting a reputational buffering effect, that is, a tendency for reviewers to adopt a more cautious or deferential tone when evaluating authors from highly prestigious institutions. In contrast, authors from less prestigious institutions were subject to greater tone variability, potentially reflecting increased scrutiny.

Together, these findings reveal clear patterns in how sentiment varies with author identity and institutional context, suggesting that reviewer tone is not applied uniformly across academic subgroups. While sentiment analysis offers a high-level view of evaluative valence, it does not capture the specific linguistic features through which tone is conveyed. To delve deeper into these patterns, we next examine the use of evaluative language in peer reviews through a detailed n-gram analysis.

\subsection*{Evaluative Language Patterns across Author Groups}

Next, we explore the kinds of language that appear most frequently in peer reviews and analyze how their use varies across author demographics and institutional characteristics (gender, race, institutional rank, and geography). To guide our analysis, we drew on established frameworks in feedback theory and scientific communication. Hyland and Diani~\cite{hyland2009introduction} describe peer reviews as a combination of praise, critique, and suggestions, while prior work highlights the role of constructive feedback in learning~\cite{Mckeachie, nicol2006formative}, appreciation in reinforcing motivation~\cite{hattie2007power}, critique in promoting scientific rigor~\cite{boud2013feedback}, and clarification in facilitating understanding~\cite{carless2006differing}. Reflecting these principles, we grouped frequently occurring bigrams and trigrams into four interpretive categories: ``manuscript appreciation,'' ``constructive suggestions,'' ``requests for clarification,'' and ``strong methodological critique.'' Each review was assessed for the presence of these expressions, with frequencies normalized per 1,000 words to account for variation in review length. For each n-gram category and demographic group, we compared the average frequency of usage. Full methodological details on n-gram extraction, category construction, and statistical comparisons are provided in the Materials and Methods section.

\begin{figure}[!h]
    \centering
    \includegraphics[width=1\textwidth]{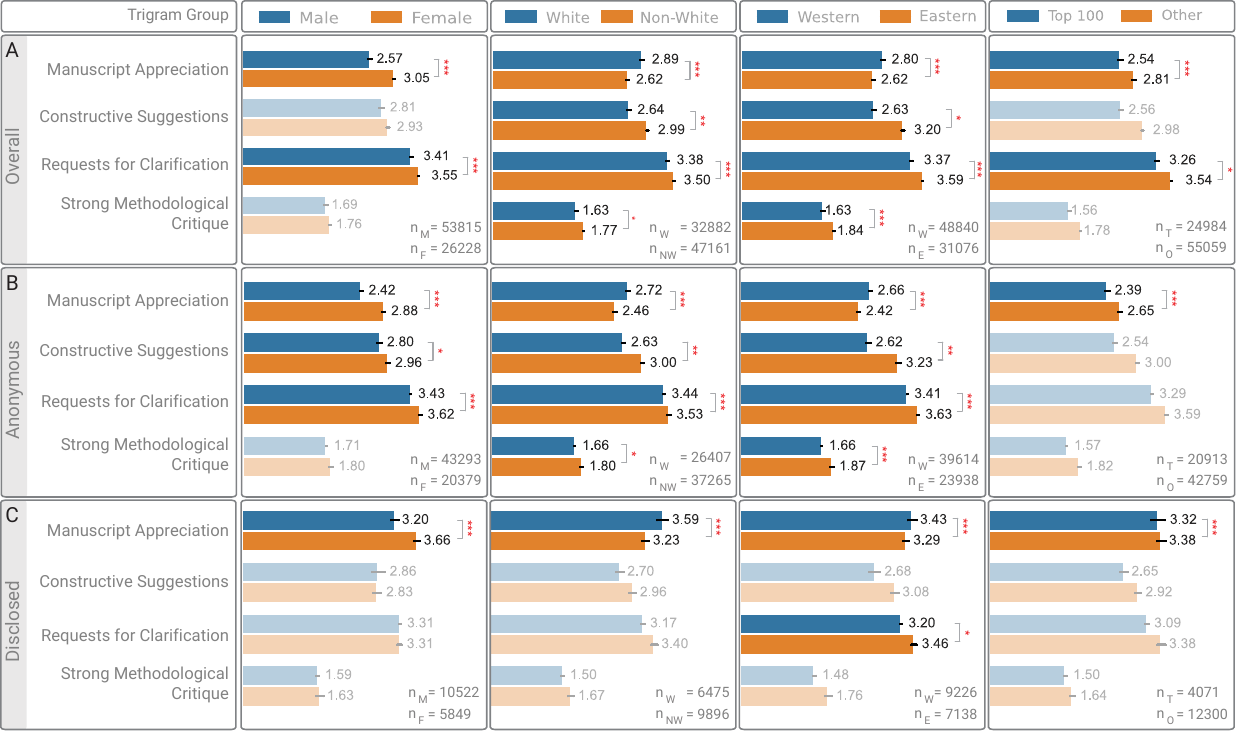}
    \caption{\textbf{Group-level differences in linguistic pattern frequencies across author demographics.}\\ Bars show the average frequency (per 1,000 words) of four grouped bigram/trigram categories  (``manuscript appreciation'', ``constructive suggestions'', ``requests for clarification'', and ``strong methodological critique'') across author gender (male/female), race (white-non-white), region (eastern/western), and institutional ranking (top 100/other). Results are presented for the \textbf{(a)} all reviews, \textbf{(b)} reviews with anonymous reviewers, and \textbf{(c)} reviews with disclosed reviewer identity. Confounder group names and color codes are indicated at the top of the figure, corresponding sample sizes for each sub-group are shown on the bottom-right side of each panel. Asterisks denote statistically significant differences between groups: *** $p < 0.001$, ** $p < 0.01$, * $p < 0.05$. Faded bars represent comparisons where the group-level difference did not reach statistical significance.}
    \label{fig:trigrams_bigrams}
\end{figure}

Figure~\ref{fig:trigrams_bigrams}a reveals a distinct linguistic pattern in reviewer feedback, with authors from marginalized or less prestigious backgrounds more often receiving correction-oriented, clarifying, or critical language. More specifically, non-white and eastern-affiliated authors received significantly more n-grams related to constructive suggestions, requests for clarification, and strong methodological critique compared to their white and western-affiliated counterparts. Female and non-top 100 authors also received significantly more clarification n-grams, and although the differences in constructiveness and critique for these groups were not statistically significant, they followed the same directional pattern.

In contrast, patterns of appreciative language were more mixed. White and Western-affiliated authors received the highest levels of appreciation overall, yet female and non-top 100 authors also received significantly more appreciation than male and top 100 counterparts. These findings indicate that while corrective and clarifying feedback tends to concentrate around authors from less privileged or non-dominant groups, affirming language is distributed in a more complex, less hierarchical pattern. 

To better understand whether these disparities are shaped by the visibility of reviewer identity, we repeated the analysis separately for reviews where the reviewer remained anonymous versus not (Figure~\ref{fig:trigrams_bigrams}b and c). Across all four demographic dimensions, group-level differences in reviewer language were consistently similar in anonymous reviews. For example, gender-based disparities in reviewer language remained statistically significant under anonymity for appreciation and clarification, and even gained significance in constructive suggestions; however, under disclosure, only the appreciation gap remained significant. Similarly, racial and geographic differences in critique, constructiveness, and clarification largely disappeared in signed reviews, while appreciation differences persisted. In the case of institutional rank, only appreciation remained significantly higher for authors from non-Top 100 institutions in disclosed reviews, while other categories showed no significant variation.

Building on prior work linking language and evaluative outcomes in peer review~\cite{falk2019language, al2022analysis, sun2024textual}, our study contributes a fine-grained n-gram analysis that uncovers systematic variation in reviewer phrasing across social and institutional lines. While these descriptive comparisons offer valuable insights into how language use varies by author demographics and reviewer anonymity, the observed differences are relatively modest in magnitude and do not account for the potential interplay between author characteristics. Furthermore, n-gram analysis captures recurring language patterns, but does not convey the tone embedded in those expressions, a crucial element of written communication, particularly in peer reviews where non-verbal cues are absent~\cite{mehrabian1971silent, picard1997w}. To more rigorously assess both what reviewers say and how they say it, we next apply large language models (LLMs) to classify tone and use regression-based analyses to examine the independent and interacting effects of author characteristics.


\subsection*{Tone Variation across Author Groups}

To build on the sentiment and trigram-level analyses, we applied multivariate regression models to examine how reviewer tone is shaped by characteristics of the corresponding author, the manuscript, and the review itself. While earlier sections identified broad trends in sentiment and language patterns, regression analysis enabled us to isolate the unique contribution of each factor and assess whether certain groups consistently receive more appreciative, constructive, questioning, or critical feedback.

Tone classification was performed at the sentence level using GPT-4o, a large language model capable of capturing both explicit and nuanced forms of evaluation. We designed a structured prompt to guide the model in assigning each sentence to one of four tone categories: appreciative, constructive-analytical, questioning, or critical-evaluative. This approach allowed for systematic and scalable tone analysis across a large corpus of peer review texts. By combining a literature-informed and domain-specific tone framework with GPT-4o’s advanced language understanding capabilities, we were able to detect both overt and subtle expressions of feedback, offering a fine-grained lens into the language of peer evaluation.

Furthermore, to validate the accuracy of the model’s classifications, we conducted two complementary validation exercises. The first involved corresponding authors of the reviewed manuscripts, yielding a high agreement rate of 91.1\%. The second involved impartial researchers with no connection to the manuscripts, who achieved an accuracy of 95.1\%. These results confirm that the tone classifications were both domain-relevant and independently reliable. Full details on the prompt design, classification procedure, and validation methods are provided in the Materials and Methods section.

For each review, we quantified the prevalence of each tone category using a weighted tone score that accounts for both the frequency and length of tone-labeled sentences. These scores served as dependent variables in a series of regression models, which included a common set of predictors across three domains: (1) demographic and institutional characteristics of the corresponding author, including gender, race, affiliation rank, region, academic age, number of prior publications, and average prior impact; (2) manuscript-level features, including publication year, review duration, number of coauthors, and journal; and (3) review-specific characteristics, such as review length. Full details on tone score construction, variable creation, and data preprocessing and diagnostic procedures are provided in the Materials and Methods section. Further details on tone score construction and variable definitions are described in the Materials and Methods section. 

To examine how tone varies by author characteristics and reviewer anonymity, we estimated three sets of regression models for each tone category: one using the full dataset, and two additional sets stratified by reviewer identity (anonymous and disclosed). In the full dataset models, reviewer identity disclosure was included as a binary predictor (1 = disclosed, 0 = anonymous). This variable showed a significant association with tone in three out of four categories. Specifically, the reviews with disclosed reviewer identity were significantly more appreciative, constructive, and questioning in tone compared to anonymous reviews (see Supplementary Table~\ref{tab:regression_results_full}). While critical tone was slightly lower in disclosed reviews, this difference was not statistically significant. These shifts suggest that when the identities of the reviewers are known, the reviewers may adopt a more engaged and respectful tone, possibly due to increased social accountability.

\begin{figure}[!h]
    \centering
    \includegraphics[width=1\textwidth]{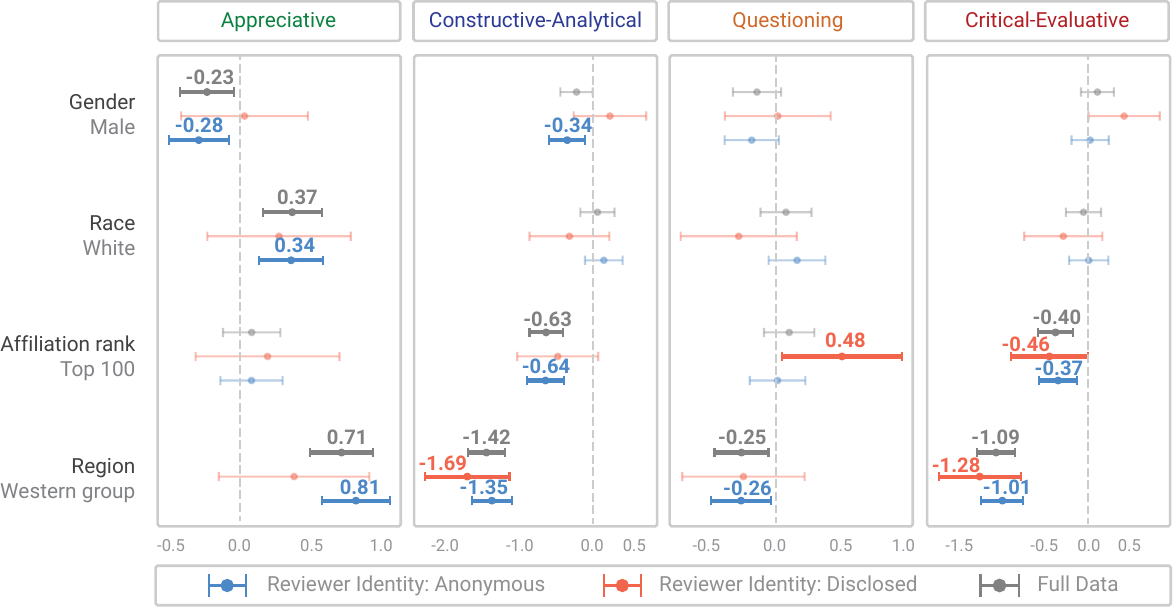}
    \caption{\textbf{Regression coefficients and $95\%$ confidence intervals estimated for each tone category using the full dataset (gray) and reviewer-stratified models (blue = anonymous, orange = disclosed).}\\ The first row displays the effect of reviewer identity disclosure, included as a binary indicator in the full dataset models only. As this variable showed a significant association with tone, we subsequently estimated separate models for disclosed and anonymous reviews. The plot presents results for four core author-level features: gender (reference: male), race (reference: non-white), institutional rank (reference: non–Top 100), and regional affiliation (reference: non-Western group). Statistically significant coefficients are annotated in the corresponding model color. The vertical dashed line denotes the null value (zero effect). }
    \label{fig:regression_summary_main}
\end{figure}

Motivated by this pattern in the full dataset, we then stratified the analysis and estimated separate models for anonymous and disclosed reviews to assess whether the relationships between author characteristics and tone differ by reviewer identity status. Supplementary Table~\ref{tab:regression_results_disclosed_anon} presents a comprehensive view of all predictors included in the models, while Figure~\ref{fig:regression_summary_main} offers a more focused summary by highlighting key author-related characteristics, namely the corresponding author’s gender, race, institutional rank, regional affiliation, and reviewer identity disclosure status.

The figure reveals that biases in tone do indeed vary depending on whether the review was anonymous or signed. Notably, the tendency for reviewers to use more appreciative language when evaluating submissions from female, white, or eastern-affiliated corresponding authors disappears entirely when reviewer identity is disclosed, suggesting that such praise may be more freely expressed under anonymity. Similarly, the use of more constructive language for non-top-100 affiliated authors vanishes under disclosure. Questioning tone toward eastern-affiliated authors also diminishes under signed review, pointing to greater caution in expressing doubt when attribution is possible.

In contrast, disparities in critical-evaluative tone persist regardless of reviewer anonymity. Authors from institutions outside the top-ranked or from eastern affiliations consistently receive more critical feedback, suggesting possible perceptions of lower credibility or quality. Constructive-analytical tone toward eastern-affiliated authors also remains consistent, suggesting that reviewers continue to offer guidance and suggestions to these authors regardless of whether their identity is known. Additionally, characteristics beyond demographics and institutional affiliation also appear to shape tone. Notably, academic age was negatively associated with all tone categories, suggesting that more senior authors tend to receive less tone-marked feedback overall, regardless of whether the review is anonymous or signed.

Interestingly, questioning tone followed a different pattern in one notable case: it became more prominent in disclosed reviews for authors affiliated with top-ranked institutions. This may reflect a shift in reviewer behavior under open conditions, where perceived prestige invites greater engagement through questions and clarification. Notably, questioning tone often reflects a desire to better understand the authors’ choices or to prompt elaboration, such as asking why a specific method was used or whether an additional analysis could strengthen the work. These forms of questioning may serve a constructive function in encouraging the authors to clarify, justify, or expand their arguments, rather than indicating disapproval.

Overall, the results show that reviewer tone is shaped by a combination of social and procedural factors. Reviewer anonymity, author characteristics, and review structure each contribute in distinct ways to how tone is expressed in peer evaluation. While the influence of author characteristics varied across tone categories, disparities were more pronounced under anonymous review, particularly for appreciative and constructive tones, whereas identity disclosure was associated with more consistent and balanced feedback. These findings highlight the complex interplay between reviewer context and author attributes in shaping the language of peer review.

To further examine how reviewer tone changes over time, we conducted an additional analysis testing interactions between reviewer identity disclosure and publication year (see Supplementary Note~\ref{note:tone_temporal} and Supplementary Figure~\ref{note:tone_temporal}). While both anonymous and disclosed reviewers showed increases in appreciative and constructive tones, the rate of increase was significantly greater among disclosed reviewers, suggesting a growing shift toward more supportive review practices when identities are known. The questioning tone, on the contrary, decreased over time in disclosed reviews, potentially reflecting a move away from interrogative language under conditions of visibility. Critical-evaluative tone remained stable across time and identity conditions, indicating that increased appreciation and constructiveness did not come at the expense of critique. Together, these findings suggest that reviewer identity disclosure not only influences tone at a given point but may also shape how the tone evolves over time.

These patterns have important implications for editorial practices. Anonymity may create conditions where tone differences based on author identity are more likely to emerge, while disclosure appears to promote greater accountability and a more constructive reviewing style. 
Together, these insights can help shape journal policies that promote fairness, reduce unintended bias, and encourage constructive, respectful, and high quality peer reviews.

\section*{Discussion}
This study reveals consistent disparities in how reviewers evaluate manuscripts, with reviewer tone systematically shaped by the demographics and institutional affiliations of corresponding authors. Female, white, and western-affiliated authors were more likely to receive appreciative and positive sentiment, especially under anonymous review. In contrast, non-white and eastern-affiliated authors were more frequently met with critical and negative language. They also received a higher proportion of constructive feedback, often framed as suggestions or improvement-oriented comments, which, while not necessarily negative, may reflect underlying assumptions about competence or credibility.

Importantly, reviewer identity disclosure was associated with a marked reduction in these disparities, particularly for appreciative tone. When reviewers signed their names, the elevated levels of appreciation observed in reviews of white, female, and western-affiliated authors largely disappeared. However, disclosure did not fully eliminate disparities in critical or constructive language. Authors from non-western or lower-prestige institutions continued to receive more critical and improvement-oriented feedback even under disclosure. That said, disclosure did not amplify these patterns either. Rather, it appeared to moderate extremes in tone, encouraging more balanced and measured expression.

Our findings suggest that, while anonymity does not necessarily exacerbate existing social biases, it also fails to mitigate them. In contrast, reviewer identity disclosure appears to foster a more balanced and equitable evaluative environment. It encourages accountability, reduces unwarranted extremes, whether in praise or criticism, and promotes a tone that is more supportive without compromising rigor. Importantly, we found no evidence that disclosure leads to harsher reviews, addressing a concern that is shared by editors and reviewers alike.

These insights carry important implications for peer review policy. While the double-anonymous peer-review model has been shown to reduce certain biases related to author identity, such as gender, institutional prestige, or geographic origin~\cite{cuskley2020double, kern2022impact}, it remains an incomplete solution. This model may help level the playing field in some respects, but it does not address issues like ghostwriting~\cite{mcdowell2019co}, nor does it fully protect work that engages with stigmatized or controversial topics from biased evaluation~\cite{strauss2023racism}. One limitation of our study is that we did not include double-anonymous reviews, as our dataset was restricted to reviews written under single-anonymous or disclosed conditions. However, this limitation does not undermine our central findings. Our results highlight the potential value of reviewer disclosure even in contexts where author identities are known. Thus, even in contexts where author anonymity is not feasible or uniformly applied, encouraging reviewer transparency may still offer a viable pathway toward greater fairness and accountability.

However, adopting full transparency is not without challenges. One major concern is reviewer recruitment, which is already facing significant difficulties. The peer review system, which has long relied on volunteer labor, has become increasingly overburdened, especially since the COVID-19 pandemic, due to the increasing volume of submissions and widespread reviewer fatigue~\cite{flaherty2022peerreview}. As more reviewers decline invitations, the burden falls on a shrinking pool, leading to delays and potentially compromising review quality~\cite{huisman2017duration, drozdz2024peer}. Limited reviewer availability also increases the risk of biased or conflicted evaluations, particularly when relationships between editors, reviewers, and authors are undisclosed~\cite{mcintosh2025safeguarding}. These structural challenges may make some scholars, especially early-career ones, hesitant to accept assignments if disclosure is expected. Nonetheless, building a culture of accountability, in which reviewers take responsibility for their words, remains an important goal for the academic community.

Senior scholars in particular have a responsibility to lead by example. Surveys show that postdoctoral researchers and graduate students often draft reviews on behalf of invited reviewers without receiving formal recognition, a practice known as ghostwriting. This not only conflicts with community values, but also violates the policies of most academic journals~\cite{mcdowell2019co}. Requiring disclosure of all contributors to a review could incentivize senior academics to more closely supervise, mentor, and take ownership of the feedback provided, while also promoting transparency and ethical co-review practices. By highlighting key patterns and trade-offs, this study helps inform broader discussions on how to improve the fairness, transparency, and accountability of peer review in ways that reflect the values of scholarly rigor and inclusivity.

\pagebreak
\section*{Materials and Methods}

\subsection*{Data}

To initiate our analysis, we gathered peer reviews for approximately 50,000 articles from two leading open-access platforms:

\begin{enumerate}
    \item \textbf{Nature Communications:} We sourced publicly available metadata and peer reviews from the Nature Communications website, ensuring strict compliance with the publisher's terms of service and data usage policies. Only openly accessible data were collected to adhere to all legal and ethical guidelines for data collection~\cite{nature_self_archiving}.

    \item \textbf{PLOS One:} We acquired data from PLOS ONE's openly accessible XML file repository. The structured and machine-readable format allowed for efficient extraction of both metadata and peer review content directly from their website. This data was collected in accordance with all legal and ethical guidelines for data collection~\cite{plos_tdm}.
\end{enumerate}

To maintain the quality and reliability of the data set, we applied several exclusion criteria. Reviews containing fewer than 200 characters or those containing only placeholders such as ``see attached files'' were excluded. Our focus was on articles published from 2019 onward, marking the first year that both journals adopted a policy that allowed reviewers to disclose their identities. Furthermore, since our study examines biases based on the characteristics of corresponding authors, we excluded manuscripts with multiple corresponding authors to avoid ambiguity in attributing decisions or biases to specific personas.

In preparation for our analysis, we also extracted and engineered various features categorized into three groups: (1) corresponding author-related features, (2) paper-related features, and (3) review-related features. In the following, we detail each set of features and the methods used to obtain them.

\begin{enumerate}
\item\textbf{Corresponding Author-Related Features:}

\begin{itemize}
    \item \textbf{Gender:}  The gender of each author was determined based on their first name using the \textit{Genderize.io} API~\cite{web:Genderize}, which provides probabilistic estimates of gender based on large-scale name databases. To assess the accuracy of this classifier, we invited corresponding authors to self-report their gender. Out of 531 survey respondents, 445 provided this information. Based on their responses, the classifier achieved an accuracy of 94.4\%.
    
    \item \textbf{Race:} To estimate the race of authors, we utilized \textit{NamSor} API~\cite{web:Namsor}, which is a software reinforced by more than 7.5 billion names processed and accurately classifies personal names by country of origin and race. While race is a complex and multi-dimensional concept, \textit{NamSor} offers a reasonable proxy for the broad ethnic categories considered in this study. It provides four main race categories: White, Black, Hispanic/Latino, and Asian. To assess the accuracy of this classifier, we contacted corresponding authors of the reviewed manuscripts and asked them to self-identify their race. Of the 531 survey respondents, 430 provided their racial identity, allowing us to evaluate the classifier’s performance. On this subset, the classifier achieved an accuracy of 83.2\%

    \item \textbf{Affiliation rank:} Given that the affiliations could range from universities to companies to research laboratories, we opted to identify the rank of each author's institution using three ranking systems: \textit{Times Higher Education}~\cite{web:TIMES}, \textit{QS World University Rankings}~\cite{web:QS}, and the \textit{Nature Index}~\cite{web:NatureIndex}. If the author’s institution was ranked in the top 100 in any of these lists, we categorized them as being affiliated with a ``Top 100'' institution. Otherwise, the institution was not flagged as a top-ranked university or company.
    
    \item \textbf{Country of affiliation:} Each author's country of affiliation was extracted from their affiliation data at the time of the paper's publication. The country data was used to classify each author’s location into Eastern or Western regions, based on geopolitical and cultural considerations. The classification adhered to established regional categorizations without overlap between Eastern and Western countries. 
    \item \textbf{Academic age:} We utilized the \textit{OpenAlex} API~\cite{openalex}, a comprehensive open-access database that compiles detailed information on scholarly papers and researchers, to enhance our dataset by matching paper titles with OpenAlex's records. This enabled us to obtain insights into the authors' publication histories and calculate the academic age of each author, defined as the number of years since their first publication recorded in the OpenAlex database.
    \item \textbf{Number of prior publications:} In addition to determining academic age, we utilized OpenAlex to collect data on each author's productivity, measured by the total number of publications up to the time of review.
    \item \textbf{Average prior impact:} We calculate the average impact of papers published by each author before the reviewed paper using the OpenAlex database. Due to the recent nature of our publications, and following common practices in the scientometrics literature, we adopted the approach suggested by AlShebli et al. 
    \cite{alshebli2022beijing, alshebli2024china}. This approach measures impact by counting the number of citations a paper receives within the first two years after publication. This time frame is widely recognized as an indicator that is highly associated with the overall citation performance over longer periods.
\end{itemize}

\item \textbf{Paper-Related Features:}

\begin{itemize}
    \item \textbf{Field of science:} Each paper was matched with the OpenAlex database. Using OpenAlex, we retrieved the primary field of science assigned to each publication. This classification allows us to analyze differences in review tone and other characteristics across 24 various academic disciplines, such as Biology, Chemistry, Physics, and Medicine, among others. The full list of fields, along with their grouping into broader scientific domains is presented in Supplementary Figure~\ref{fig:suppl_fields_of_science}.
    \item \textbf{Publication year:} We recorded the publication year for each paper to analyze trends in peer review over time and observe shifts in reviewer tone influenced by evolving editorial practices, standards, and global events. Our dataset focuses on papers published from 2019 to 2024, aligning with the policies of Nature Communications and PLOS ONE, which introduced transparent peer review options in 2016 and 2019, respectively, allowing reviewers' identities and comments to be disclosed. Consequently, our analysis begins in 2019.
    \item \textbf{Review duration:} For each paper, we recorded the submission date and the acceptance date. These dates are important for understanding how long the review process takes, from initial submission to acceptance for publication. By calculating the number of days between submission and acceptance, we assessed the speed and efficiency of the review process. 
    \item \textbf{Team size:} For each paper, we extracted the number of co-authors on the manuscript.
    \item \textbf{Journal:} Each paper, and therefore review, in the dataset was published in either Nature Communications or PLOS ONE. This variable allows us to control for any differences in review processes or editorial standards between the two journals.
\end{itemize}

\item \textbf{Review-Related Features:}

\begin{itemize}
    \item \textbf{Review length:} For each paper, we identify the length of the review by counting the number of words in it. The length of a review reflects the level of effort, detail, or engagement a reviewer puts into their feedback. Longer reviews may indicate greater thoroughness or a stronger commitment to the review process, which could correlate with differences in tone, criticism, or the type of feedback provided. 
    \item \textbf{Reviewer anonymity:} We identify whether or not the reviewer chose to disclose their identity. 
\end{itemize}
\end{enumerate}

Finally, after extracting our features, we excluded papers if the corresponding author's (i) gender or race could not be determined due to incomplete names or if the confidence level for identifying gender or race was below 70\%, or if (ii) the institution name or regional affiliation could not be reliably extracted from the metadata. Furthermore, papers not found in the OpenAlex database were excluded, as we required information such as academic age and the number of publications of the corresponding author. This resulted in a final dataset of 32,862 academic papers and their corresponding peer reviews. A summary of the characteristics of the dataset can be found in Table~\ref{tab:datadescription}.

\begin{table*}[!ht]
\centering
{\fontsize{12}{12}\selectfont{
\renewcommand{\arraystretch}{1.5}
\begin{tabular}{|>{\raggedright\arraybackslash}p{4.25cm}|>{\raggedright\arraybackslash}p{2cm}|>{\raggedright\arraybackslash}p{2cm}|>{\raggedright\arraybackslash}p{2cm}|>{\raggedright\arraybackslash}p{2.5cm}|}
\hline
Journal & Time frame & Number of manuscripts & Number of reviews & Number of signed reviews\\
\hline 
Nature Communications & 2019-2024 & 8,292 & 19,462 & 1,126\\
PLOS ONE & 2019-2024 & 24,570 & 60,725 & 15,410\\
\hline
\end{tabular}
}}
\caption{\textbf{Final dataset.} Summary of the peer review data used in this analysis, including the number of articles, peer reviews, and signed reviews from Nature Communications and PLOS ONE (2019–2024).}\label{tab:datadescription}
\vspace{-4mm}
\end{table*}

\subsection*{Using SciBERT to classify sentiment}

To perform sentiment analysis, we selected SciBERT~\cite{beltagy2019scibert}, a transformer-based language model adapted from BERT~\cite{devlin2018bert} and pretrained specifically on scientific texts. Unlike general-purpose language models, SciBERT was trained on more than 1 million full-text scientific papers and 3 million abstracts from the Semantic Scholar corpus. This training makes it especially suitable for handling the technical vocabulary and structure commonly found in peer reviews. However, while SciBERT is well aligned with the scientific domain, it is not designed for sentiment classification by default. To make it effective for this task, we fine-tuned the model using a labeled dataset of peer review texts.
  
For fine-tuning, we used the Peer Review Analyze dataset~\cite{ghosal2022peer}, which contains 1,199 open peer reviews annotated at the sentence level. The dataset includes multiple annotation layers, but for this task we focused only on the sentiment labels: \textit{positive}, \textit{neutral}, and \textit{negative}. After filtering out entries with missing sentiment labels, the final dataset consisted of approximately 17,000 labeled sentences, providing a suitable foundation for training a domain-specific sentiment classifier. 
 
During training, we used a learning rate of $2 \times 10^{-5}$, batch size of 16, weight decay of 0.01, and trained for a maximum of 10 epochs, with early stopping triggered if validation performance did not improve for two consecutive epochs. We monitored both accuracy and weighted F1-score to evaluate classification performance. The best-performing model achieved a validation accuracy of 0.677 and a weighted F1 score of 0.671. After fine-tuning, the model was used to classify the sentiment of our full set of peer reviews.

\subsection*{Identifying and Mapping N-grams in Reviewer Language}

To analyze language patterns in peer review texts, we performed an n-gram analysis of the entire corpus of reviews. N-grams defined as continuous sequences of n words, were extracted at the levels of bigram ($n = 2$), trigram ($n = 3$) and four-gram ($n = 4$). Bigrams often lacked sufficient contextual precision (e.g., “main, findings”), while four-grams were too sparse to yield reliable insights. Trigrams offered the ideal balance: they were frequent enough to be statistically meaningful and semantically rich, capturing expressions such as “not, enough, detail” and “section, well, described.”

Based on this evaluation, we adopted a hybrid approach to detect evaluative language, prioritizing trigrams for their contextual depth while incorporating a carefully curated subset of bigrams that conveyed clear evaluative signals (e.g., ``well written'', ``please clarify'', ``thank you''). This strategy provided broad coverage while preserving interpretability, allowing us to identify both nuanced and overt expressions of reviewer stance.

To implement this approach, we first preprocessed the review texts. Each review was segmented into sentences using a custom tokenizer tailored for scientific writing. This tokenizer accounted for domain-specific abbreviations (e.g., et al., Dr., Fig., i.e., e.g.) and numerical expressions (e.g., 1.2, 19.4 mm/yr) that commonly confound generic sentence splitters. The sentences were then tokenized, lowercased, and eliminated from punctuation. We remove standard English stopwords and extend the list with domain-specific terms (e.g., ``doi'', ``plosone'', ``nature'', ``communications'', ``reviewer'', ``submission''). However, we retained functional words such as negators (``not'', ``no''), modals (``should'', ``could'', ``may''), and intensifiers (``very'', ``more''), which are critical to evaluative expression. Tokens with two or fewer characters were excluded unless they carried meaningful semantic value.

Next, we extracted our n-grams using a non-overlapping sliding window and applied a variable-length n-gram matching strategy. To construct a reliable lexicon of evaluative expressions, we generated the full set of trigrams and bigrams from the corpus, approximately 4.6 million unique trigrams and 3.5 million bigrams. We then applied frequency-based filtering, retaining only trigrams that appeared more than 100 times and bigrams more than 1,000 times. As such, bigrams were used more selectively and were included only if they carried strong evaluative content (e.g., ``well written'', ``please clarify'', ``please add''). This filtering process yielded a final lexicon of 269 n-grams.

Each retained n-gram was manually categorized into one of five reviewer stance categories:
\begin{enumerate}
    \item \textbf{Manuscript Appreciation} (e.g., ``well, written, manuscript''),
    \item \textbf{Requests for Clarification} (e.g., ``provide, more, detailed''),
    \item \textbf{Constructive Suggestions} (e.g., ``manuscript, could, benefit''), and
    \item \textbf{Strong Methodological Critique} (e.g., ``main, concern, with''),
\end{enumerate} 

N-grams that were procedural or ambiguous in the evaluative content (e.g., ``factors, associated, with'') were excluded. All extracted bigrams and trigrams from each category are listed in Supplementary Table~\ref{tab:suppl_trigram_categories}. Furthermore, to account for variation in review length, we normalized all n-gram counts by the number of preprocessed tokens per review, yielding frequencies per 1,000 words. This enabled consistent and interpretable comparisons of the evaluative language across reviews of varying lengths. 

We then mapped each retained n-gram to individual reviews and analyzed their distribution across four author-related axes:
\begin{enumerate}
    \item \textbf{Race} (white vs. non-white),
    \item \textbf{Gender} (male vs. female),
    \item \textbf{Institutional Geography} (Western vs. Eastern affiliation), and
    \item \textbf{Institutional Rank} (top 100 vs. other)
\end{enumerate}

For each evaluative category and demographic group, we computed the average number of n-gram matches per 1,000 words per review, allowing for direct comparisons while accounting for differences in sample size. To assess whether language pattern frequencies differed significantly across groups, we applied the non-parametric Mann-Whitney U test, which compares rank distributions between two independent samples without assuming normality. To confirm the robustness of these results, we performed bootstrap resampling with 1,000 iterations, which consistently supported the significance of group differences, despite the small effect sizes observed. Finally, as an additional validation step, we used the two-sample Kolmogorov-Smirnov test to determine whether the overall distributions of n-gram frequencies differed significantly between groups and the overall results were consistent.


\subsection*{Tone Analysis}

\subsubsection*{Tone classification}
To systematically analyze tone in peer review texts, we used OpenAI's GPT-4o API with a structured prompt to classify each sentence into one of four tone categories: ``Appreciative'', ``Constructive-Analytical'', ``Questioning'', and ``Critical-Evaluative''. More specifically, the four tone categories were defined as follows:

\begin{itemize}

    \item \textbf{Appreciative:} Expresses approval, recognition, or gratitude for the manuscript’s contribution or clarity. Example: \textit{``This study makes a significant contribution to the field. Your clear presentation and thorough analysis are commendable.''}
    
    \item \textbf{Constructive-Analytical:} Provides a detailed examination of methodology, data, or results, offering targeted suggestions to enhance the research. Example: \textit{``The methodology is robust, but expanding your sample size could improve the generalizability of your findings. Including a sensitivity analysis would also strengthen your conclusions.''}
    
    \item \textbf{Questioning:} Seeks clarification or elaboration, often indicating uncertainty or lack of information. Example: \textit{``Can you provide more detail on the selection criteria for your sample population? It is not entirely clear how the inclusion and exclusion criteria were determined.''}

    \item \textbf{Critical-Evaluative:} Highlights weaknesses or limitations without offering solutions, focusing on shortcomings in methods, claims, or results. Example: \textit{``Statistical analysis does not adequately support the conclusions. The small sample size limits the study's validity and reliability.''}

\end{itemize}

Furthermore, and in line with peer review guidelines~\cite{web:NatureCommGuidelines, web:PlosOneGuidelines}, which recommend separating summaries from critique, we also included a fifth category, ``Contextual-Summarizing'' to label descriptive or neutral sentences that introduce or summarize the manuscript without evaluative content. These were included in the total sentence counts but excluded from tone-based analysis. 

As such, each peer review was submitted to the model using the same structured prompt, which guided the model through several steps: (1) reading the review in full for context, (2) segmenting it into sentences, (3) assigning each sentence a tone label based on defined categories, (4) identifying any contextual-summarizing sentences (5) preserving sentence order, (6) identifying signature lines, and (7) extracting reviewer names or affiliations when present. The resulting output, structured in JSON format, enabled robust quantitative analyses of tone patterns across reviewer and author groups. The full prompt text and sample output are provided in Supplementary Note~\ref{note:prompt}.

\subsubsection*{Validation of Tone Classification}

To assess the reliability of our tone classification, we performed two complementary validation exercises. First, we asked three independent researchers, who have a track record of publishing in peer-reviewed journals and experience reading peer reviews, but were not connected to the manuscripts in question, to evaluate whether the LLM tone labels accurately reflected the content and intent of each sentence in a given review. This form of external validation ensures an impartial assessment of tone classification quality, grounded in general scientific literacy and peer review familiarity. Second, we invited scientists whose manuscripts were reviewed to evaluate the LLM tone labels on their own reviews. While authors may have subjective biases, particularly in how they interpret critical or appreciative feedback about their work, they also possess the most relevant domain expertise and are best positioned to recognize field-specific nuances, jargon, and evaluative context. Together, these two validation streams allow us to triangulate classification accuracy by balancing impartiality with subject-matter expertise.

In the first validation exercise, three independent researchers evaluated 100 randomly selected peer reviews that had been annotated by the LLM using the predefined tone categories. Each sentence was assessed individually, with validators indicating whether they agreed or disagreed with the assigned tone. A sentence was considered accurately classified if at least two of the three reviewers agreed with the label. In total, 537 sentences were evaluated, yielding an overall accuracy of 95.1\%. When applying a stricter criterion requiring unanimous agreement among all three validators, the accuracy was 93.8\%.

In the second validation exercise, our goal was to collect at least 500 responses from corresponding authors in our dataset. We contacted them in batches until we reached this target, ultimately receiving 531 completed responses, corresponding to an 8\% response rate. These 531 authors collectively evaluated 1,192 peer reviews, comprising a total of 21,150 sentences, as most manuscripts included two to three reviews each. Authors were shown their own peer reviews and asked to assess whether the LLM's tone classification for each sentence was accurate. Comparing author assessments with the LLM’s classifications, we found an accuracy rate of 91.1\%; indicating that 91.1\% of sentences were correctly classified according to the authors of the reviewed manuscripts. The sample of corresponding authors was broadly representative of the full dataset, as shown in Supplementary Table~\ref{tab:confounder_distribution}, which compares key characteristics across both groups.

At the end of the survey, participants were invited to voluntarily provide demographic information, including gender and race. This optional data allowed us to validate the performance of our demographic classifiers, as reported in the Data subsection. All responses were anonymized and used solely for academic research, under the oversight and approval of our Institutional Review Board (IRB, \#HRPP-2025-83). The complete survey structure, along with an illustrated example of the interface and question format, is available in the Supplementary Note~\ref{note:survey_content}. Together, these two validation approaches ensured that our tone classification framework was grounded in both external expert consensus and the lived experience of the authors who received peer feedback.

\subsubsection*{Weighted Tone Scoring Method}
To quantify the prevalence of each tone within peer reviews, we developed a weighted tone scoring method that combines both the frequency of tone-labeled sentences and their length in words. While a simple count of tone-labeled sentences provides a useful baseline, it fails to account for the depth or elaboration of the tone expressed. For example, a brief appreciative sentence like “Well written” and a detailed paragraph elaborating on the manuscript's strengths would be weighted equally under a sentence-only scheme.

To address this, we implemented a hybrid weighting formula:

\begin{equation}
\text{Weighted}_{\text{tone}} = 100 \times \left( \alpha \cdot \frac{\text{Words}_{\text{tone}}}{\text{Total Words}} + (1 - \alpha) \cdot \frac{\text{Sentences}_{\text{tone}}}{\text{Total Sentences}} \right)
\end{equation}

where $\alpha = 0.2$ in our analysis. This weighting assigns 80\% importance to the sentence-level tone frequency and 20\% to the proportion of words, striking a balance between tone presence and elaboration. In our judgment, any value of $\alpha$ between 0.1 and 0.4 would have been reasonable, reflecting modest emphasis on length without diminishing the importance of shorter but impactful tone expressions.

This approach ensures that brief but potent comments such as “excellent work” are not overlooked, while also giving due weight to longer, more substantive evaluations. For a concrete example illustrating how this weighting method captures qualitative differences between reviews with similar sentence-level tone frequencies but different expression depth, see Supplementary Section~\ref{note:sentence_word_length}.

\pagebreak
\bibliography{bibliography}
\bibliographystyle{naturemag}
\pagebreak
\section*{Main Figures}

\clearpage

\newcommand{\beginsupplement}{
    \setcounter{section}{0}
    \renewcommand{\thesection}{\arabic{section}}
    \setcounter{equation}{0}
    \renewcommand{\theequation}{\arabic{equation}}
    \setcounter{table}{0}
    \renewcommand{\thetable}{\arabic{table}}
    \setcounter{figure}{0}
    \renewcommand{\thefigure}{\arabic{figure}} 
    \captionsetup[figure]{name=Supplementary Figure} 
    \captionsetup[table]{name=Supplementary Table}
}

\beginsupplement

\section*{\centering Supplementary Materials for Disparities in Peer Review Tone and the Role of Reviewer Anonymity}

\subsection*{Supplementary Figures}

\begin{figure}[!ht]
    \centering
    \includegraphics[width=1\textwidth]{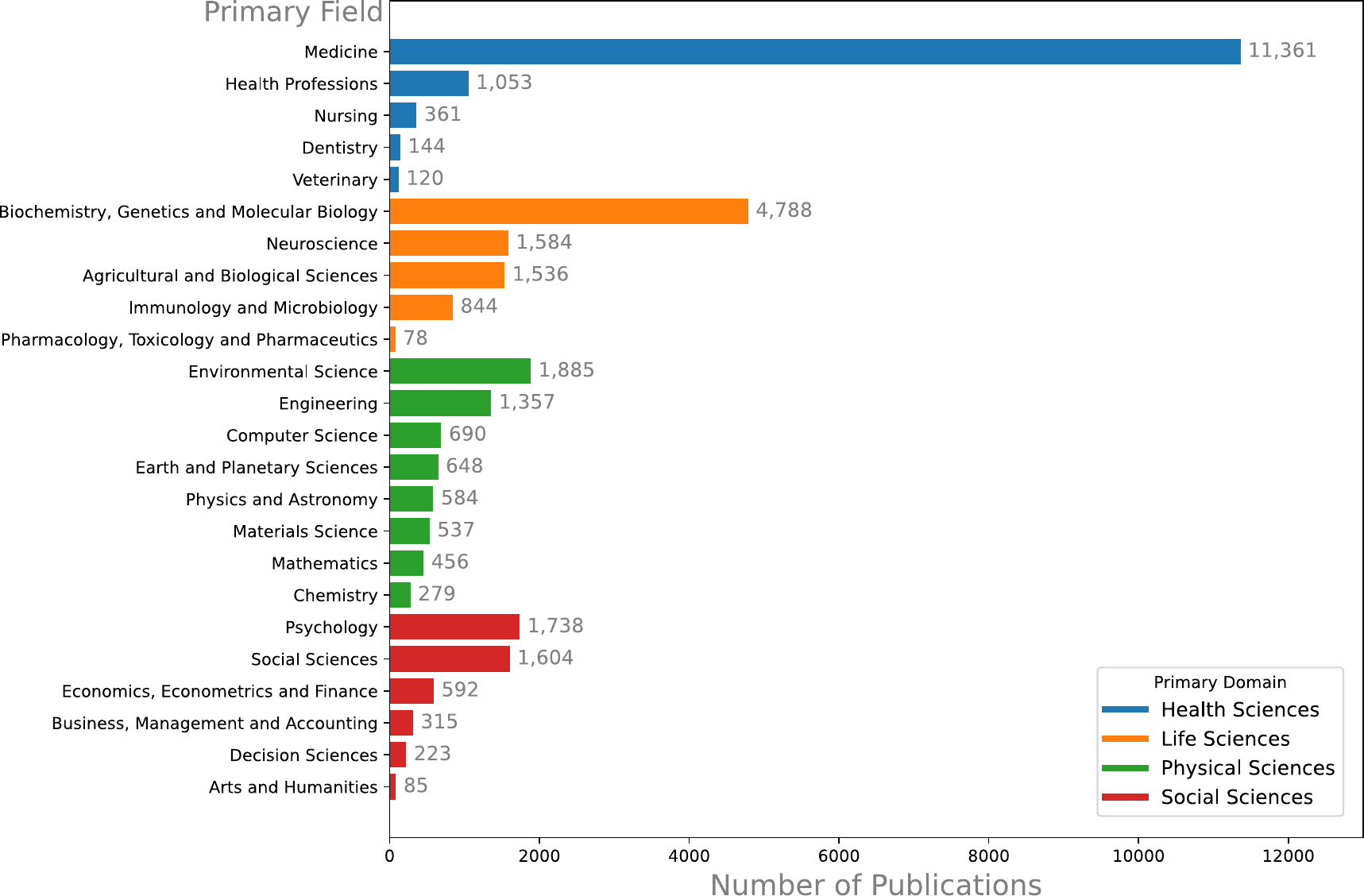}
    \caption{\textbf{Distribution of publications by primary field of science.}
    The bar plot shows the number of peer-reviewed manuscripts in the dataset categorized by their primary field, based on OpenAlex classifications. Fields are color-coded by broader scientific domain: Health Sciences (blue), Life Sciences (orange), Physical Sciences (green), and Social Sciences (red). Medicine is the most represented field and was treated as the reference category in regression analyses.
    }
    \label{fig:suppl_fields_of_science}
\end{figure}


\begin{figure}[!ht]
    \centering
    \includegraphics[width=1\textwidth]{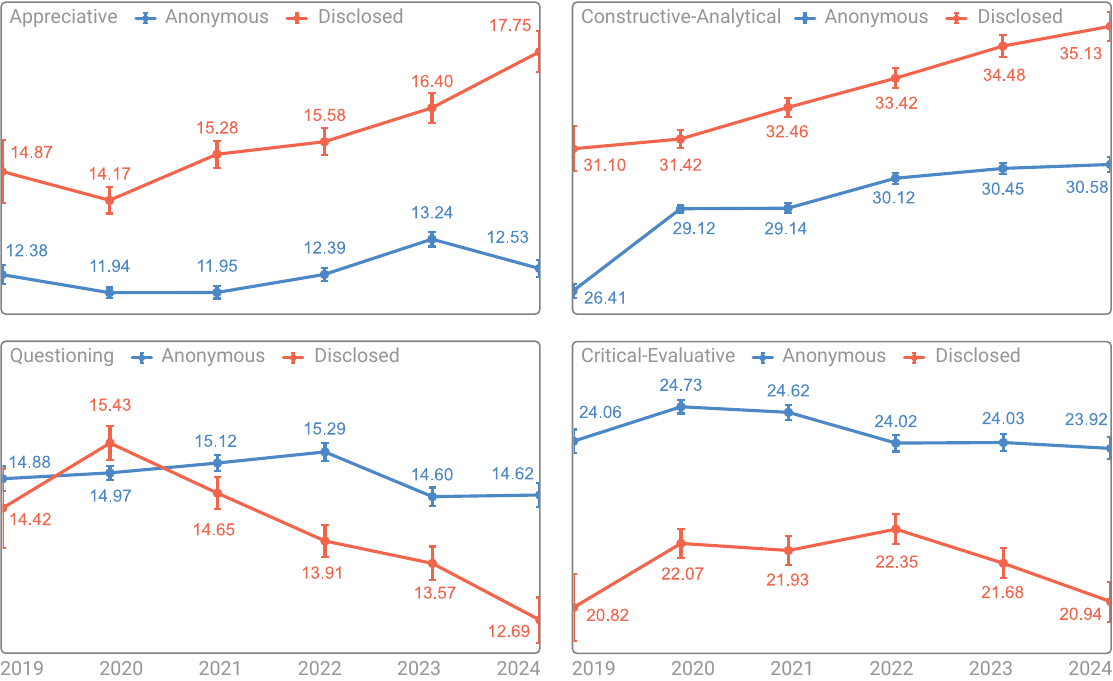}
    \caption{\textbf{Mean weighted tone scores over time by reviewer identity (Anonymous vs. Disclosed) for four tone categories.}
    Each subplot displays the annual mean scores for a given tone, with standard error bars, across publication years corresponding to 2019-2024. Disclosed reviews show a marked increase over time in both Appreciative and Constructive-Analytical tone scores compared to Anonymous reviews. In contrast, Questioning tone decreases more sharply in Disclosed reviews. Critical-Evaluative tone trends remain relatively stable over time with no clear divergence by reviewer identity. These trends reflect systematic shifts in tone expression associated with identity disclosure.
}
    \label{fig:suppl_tone_over_time}
\end{figure}


\clearpage

\subsection*{Supplementary Tables}

\begin{table*}[!ht]
\centering
{\fontsize{12}{12}\selectfont{
\renewcommand{\arraystretch}{1.5}
\begin{tabular}{|>{\raggedright\arraybackslash}p{1.5cm}|>{\raggedright\arraybackslash}p{2cm}|>
{\raggedright\arraybackslash}p{1.5cm}|>
{\raggedright\arraybackslash}p{1.5cm}|>{\raggedright\arraybackslash}p{1.5cm}|>
{\raggedright\arraybackslash}p{1.5cm}|>{\raggedright\arraybackslash}p{1.5cm}|}
\hline
 Gender & Race & Region & Ranking & Age bin [0, 10] &  Age bin [11, 25] &  Age bin [26 +)\\
\hline

 Male & Non-white & Eastern  &  Other & 6166 & 6822  & 2956 \\
 Male & White & Western & Other  & 3230  & 5159  & 3358 \\
 Male & Non-white & Western & Other & 2132 & 3526 & 1896 \\
 Female & White & Western & Other & 2837  & 2799  & 1068 \\
 Male & White & Western & Top 100 & 1611  & 2974 & 2089 \\
 Female & Non-white & Eastern & Other & 2605  & 2454 & 737\\
 Male & Non-white & Eastern & Top 100 & 674  & 2727  & 1389  \\
 Male & Non-white & Western & Top 100 & 1198  & 2336  & 1251 \\
 Female & Non-white & Western & Other & 1680  & 1910 & 776 \\
 Female & White & Western & Top 100 & 1460  & 1809  & 786 \\
 Female & Non-white & Western & Top 100 & 880 & 1295  & 467 \\
 Male & White & Eastern & Other & 751 & 910 & 249 \\
 Female & Non-white & Eastern & Top 100 & 353  & 869  & 273  \\
 Female & White & Eastern & Other & 468 & 427  & 102  \\
 Male & White & Eastern & Top 100 & 43  &170  & 114 \\
 Feale & White & Eastern & Top 100 & 26  &74  & 35 \\
\hline
\end{tabular}
}}
\caption{\textbf{Subgroup-level total review counts across academic age bins.}  Each row corresponds to a specific subgroup of corresponding authors defined by gender, race, region, and institutional rank. Total review counts represent the number of peer reviews received by papers authored by that subgroup.}\label{tab:subgroup_sizes_sentiment}
\vspace{-4mm}
\end{table*}

\begin{table}[H]
\centering
\small
\begin{tabular}{|>{\setstretch{0.6} \raggedright\arraybackslash}p{2cm}|>{\setstretch{0.45}\justifying\arraybackslash}p{13cm}|}
\hline
\textbf{\scriptsize Category} & \textbf{\scriptsize Bigrams and Trigrams} \\
\hline
{\scriptsize \textbf{Manuscript Appreciation}} & 
{\scriptsize  \noindent 
(well, written),
(technically, sound),
(manuscript, well, written), 
(good, job),
(well, 'organized),
        (well, designed),
        (nicely, done), 
        (manuscript, well),
        (paper, well), 
        (study, well), 
        (well, described), 
        (well, done), 
        (article, well), 
        (great, interest), 
        (original, approach), 
        (appreciate, work),
        (excellent, paper), 
        (excellent, work),
        (excellent, manuscript), 
        (excellent, job), 
        (paper, well, written), 
        (enjoyed, reading, manuscript),
        (authors, should, commended),
        (very, well, written), 
        (manuscript, technically, sound), 
        (article, well, written), 
        (well, written, manuscript), 
        (study, well, designed), 
        (manuscript, clearly, written),
        (paper, very, well), 
        (very, interesting, study), 
        (very, interesting, paper), 
        (well, written, paper),
        (authors, good, job), 
        (read, with, great), 
        (experiments, well, designed), 
        (study, very, interesting),
        (study, well, written),
        (study, well, conducted), 
        (enjoyed, reading, paper), 
        (article, presented, intelligible)} 
\\
\hline
{\scriptsize \textbf{Constructive Suggestions}} 
 & 
{\scriptsize \noindent (should, consider),
    (should, add),
    (should, explain),
    (please, add),
    (need, provide),
    (need, address),
    (could, consider),
    (please, provide),
    (should, include),
    (could, considered),
    (suggest, adding),
    (consider, adding),
    (please, consider),
    (could, improved),
    (suggest, add),
    (recommend, present),
    (additional, experiments),
    (better, write),
    (better, change),
    (could, help),
    (may, want),
    (may, consider),
    (may, help), 
    (lines, please, provide),
    (line, please, rewrite),
    (line, please, correct),
    (line, please, define),
    (line, replace, with),
    (line, please, include),
    (please, explain, more),
    (please, provide, data),
    (line, please, give),
    (however, authors, need),
    (authors, should, elaborate),
    (please, double, check),
    (authors, should, cite),
    (figure, authors, show),
    (suggest, authors, add),
    (authors, can, add),
    (authors, could, add),
    (helpful, authors, could),
    (strongly, recommend, authors),
    (authors, could, provide),
    (authors, should, justify),
    (strongly, suggest, authors),
    (authors, should, take),
    (authors, should, test),
    (suggest, authors, consider),
    (authors, should, carefully),
    (suggest, authors, should),
    (should, replaced, with),
    (suggest, authors, provide),
    (however, manuscript, needs),
    (line, please, consider),
    (authors, should, consider),
    (authors, should, provide),
    (authors, should, discuss),
    (authors, should, explain),
    (authors, should, include),
    (authors, should, add),
    (please, provide, more),
    (authors, may, want),
    (authors, should, show),
    (line, authors, should),
    (authors, should, address),
    (authors, need, provide),
    (authors, may, consider),
    (line, please, add),
    (authors, should, use),
    (authors, should, mention),
    (authors, should, describe),
    (authors, should, make),
    (however, authors, should),
    (authors, could, consider),
    (authors, should, more),
    (please, make, sure),
    (authors, might, want),
    (authors, should, state),
    (authors, might, consider),
    (line, please, change),
    (authors, should, check),
    (authors, need, explain),
    (please, add, more),
    (please, add, information),
    (line, should, read),
    (authors, may, wish),
    (figure, authors, should),
    (line, please, explain),
    (authors, should, perform),
    (please, provide, information),
    (discussion, authors, should),
    (authors, should, give),
    (authors, should, better),
    (authors, should, report),
    (authors, should, present),
    (authors, should, indicate),
    (authors, should, specify),
    (authors, need, address),
    (authors, should, compare),
    (authors, should, revise),
    (authors, need, show),
    (introduction, authors, should),
    (authors, need, discuss),
    (addition, authors, should),
    (can, authors, show),
    (only, few, suggestions) 
}
\\
\hline
{\scriptsize \textbf{Requests for Clarification}} & 
{\scriptsize \noindent 
(not, clear),
    (please, clarify),
    (not, sure),
    (can, authors),
    (could, authors),
    (please, explain),
    (not, understand),
    (unclear, why),
    (why, not),
    (make, clear),
    (more, clearly),
    (not, clearly),
    (explain, why),
    (more, clear),
    (why, use),
    (dont, understand),
    (why, choose),
    (could, clarify),
    (not, clear, why),
    (authors, should, clarify),
    (not, clear, whether),
    (line, not, clear),
    (however, not, clear),
    (could, not, find),
    (please, explain, why),
    (why, authors, not),
    (why, not, use),
    (can, authors, explain),
    (line, not, sure),
    (not, clear, authors),
    (can, authors, provide),
    (could, authors, provide),
    (not, clear, which),
    (not, very, clear),
    (could, authors, please),
    (can, authors, please),
    (line, please, clarify),
    (methods, not, clear),
    (figure, not, clear),
    (authors, should, clearly),
    (could, authors, explain),
    (authors, need, clarify),
    (why, authors, choose),
    (can, authors, clarify),
    (authors, not, mention),
    (not, sure, why),
    (provide, more, information),
    (please, provide, reference),
    (not, understand, why),
    (please, make, clear),
    (could, authors, clarify),
    (should, made, clear),
    (lines, not, clear),
    (introduction, page, line),
    (page, line, change),
    (not, sure, whether),
    (should, clearly, stated),
    (line, authors, mention),
    (provide, more, details),
    (lines, authors, state),
    (please, provide, details),
    (page, line, should),
    (line, please, specify),
    (page, lines, authors),
    (discussion, page, line),
    (unclear, why, authors),
    (line, authors, mean),
    (help, reader, understand),
    (please, clarify, whether),
    (sentence, not, clear),
    (not, sure, understand),
    (table, not, clear),
    (sample, size, not),
    (not, sure, authors),
    (not, clearly, stated),
    (could, provide, more),
    (line, can, authors),
    (please, clarify, why) 
}
\\
\hline
{\scriptsize \textbf{Strong Methodological Critiques}} & 
{\scriptsize \noindent 
 (not, enough),
 (not, shown), 
 (small, sample) ,
            (major, issues),
            (not, convinced),
            (major, concerns),
            (not, convincing), 
            (not, provide),
           (difficult, understand), 
           (not, sufficient),
            (major, concern),
            (major, issue),  
            (many, issues), 
            (many, concerns), 
            (not, consistent), 
            (major, limitations), 
            (major, limitation),
            (main, concern),
             (difficult, follow),
  (difficult, read),
         (doesnt, make, sense),
         (not, consistent, with),
        (data, not, shown), 
        (data, not, support),      
        (however, major, concerns), 
        (however, concerns, about),
        (sample, size, small),
        (statistical, analysis, not),
         (however, several, major),
        (however, not, convinced),
        (makes, no, sense), 
     (data, not, convincing),
 (many, grammatical, errors) ,   
 (several, major, concerns), 
        (main, concern, with),
     (not, make, sense), 
     (major, limitation, study),
    (major, concern, with), 
    (major, issues, with), 
    (major, issue, with),
    (small, sample, size), 
    (doesnt, make, sense), 
    (not, supported, data)} 
\\
\hline
\end{tabular}

\caption{\normalsize Complete list of 269 bigrams and trigrams grouped into four interpretive categories. }
\label{tab:suppl_trigram_categories}
\end{table}

\pagebreak

\begin{table}[ht]\centering
\scriptsize
\sisetup{
  table-number-alignment = center,
  table-space-text-pre ={(},
  table-space-text-post={\textsuperscript{***}},
  input-open-uncertainty={[},
  input-close-uncertainty={]},
  table-align-text-pre=false,
  table-align-text-post=false
}
\captionsetup{width=1\textwidth, justification=justified, singlelinecheck=false}

\caption{\textbf{Regression results by tone for full dataset}. Results show the estimated effects of corresponding author-related, paper-related, and review-related characteristics on the weighted proportion of each tone category (\textit{appreciative}, \textit{constructive-analytical}, \textit{questioning}, \textit{critical-evaluative}) used in peer reviews. Models were estimated using ordinary least squares (OLS) regression with Huber-White robust standard errors to account for heteroscedasticity. Variance inflation factors (VIFs) indicated no multicollinearity concerns (all VIFs $<$ 6). Reviewer identity disclosure was included as a binary predictor (1 = disclosed, 0 = anonymous), and field of science was included as a fixed effect (24-category classification).}
\label{tab:regression_results_full}
\begin{tabular}{>{\raggedleft\arraybackslash}p{4.2cm} c c c c}

\toprule
 & \textbf{Appreciative} & \textbf{Constructive-Analytical} & \textbf{Questioning} & \textbf{Critical-Evaluative}\\ 
\midrule
\textbf{Corresponding Author Features} & & & & \\
Gender: Male        & -0.227\textsuperscript{*} & -0.215 & -0.138 & 0.108 \\
                      & (0.097) & (0.110) & (0.092) & (0.099) \\
Race: White           &   0.368\textsuperscript{***}       &   0.075      &     0.084    &   -0.046       \\
                      &  (0.104)       &   (0.116)      &    (0.097)     &  (0.105)        \\
Affiliation rank: Top 100 & 0.066    &     -0.634\textsuperscript{***}      &    0.096     &    -0.398 \textsuperscript{***}    \\
                          & (0.101)    &   (0.113)      &   (0.096)      &   (0.103)       \\
Region: Western group     &  0.706\textsuperscript{***}   &   -1.423\textsuperscript{***}  &  -0.255\textsuperscript{*}   &   -1.089\textsuperscript{***}      \\
                            &  (0.110)    &   (0.124)  &    (0.103)   &  (0.112)    \\
Academic age              &  -0.041\textsuperscript{***}    &     -0.058\textsuperscript{***}     &   -0.048\textsuperscript{***}    &   -0.057\textsuperscript{***}       \\
                          &  (0.005)   &    (0.006)     &    (0.005)     &   (0.005)       \\
Number of prior publications &0.000 &   -0.000      &     0.002\textsuperscript{***}    &     0.001     \\
                          &  (0.001)   &   (0.001)      &  (0.001)       &     (0.001)    \\

Average prior impact      &   -0.004   &    0.001     &    -0.004      &     -0.006    \\
                          &  (0.003)   &     (0.003)    &   (0.003)      &     (0.003)    \\
\midrule
\textbf{Paper Features} & & & & \\
Publication year     &  0.258\textsuperscript{***}    &   0.444\textsuperscript{***}    &   0.095\textsuperscript{**}     &     -0.004   \\
                          &   (0.031) &  (0.035)        &   (0.029)     &   (0.032)       \\
Review duration       &  -0.000   &  0.003\textsuperscript{***}      &   0.003\textsuperscript{***}     &  0.009\textsuperscript{***}       \\
                       &  (0.000)  &     (0.000)   &  (0.000)      &   (0.000)    \\
Team size    & -0.034\textsuperscript{**}     &  -0.041\textsuperscript{**}    &    0.026\textsuperscript{*}    &  -0.053\textsuperscript{***}       \\
       & (0.012)   &    (0.014)     &     (0.012)   &    (0.013)   \\
Journal: Nature Communication  &   -1.585\textsuperscript{***}   &   -6.756\textsuperscript{***}     &    -3.230\textsuperscript{***}    &   -3.597\textsuperscript{***}      \\
                               &  (0.139)  &   (0.152)      &   (0.133)     &  (0.144)      \\
\midrule
\textbf{Review Features} & & & & \\

Reviewer identity disclosed      &  2.107\textsuperscript{***}    &   1.659\textsuperscript{***}     &  0.337\textsuperscript{***}      &  -0.152       \\
                                   &  (0.119)  &   (0.132)       &  (0.108)       &    (0.116)     \\ 
Review length (words)    &   -0.009\textsuperscript{***}   &   0.001\textsuperscript{***}   &  0.002\textsuperscript{***}      &   -0.001\textsuperscript{***}      \\
                       &  (0.000)  &   (0.000)      &  (0.000)      &  (0.000)     \\

Appreciative Percentage      &     &   -0.577\textsuperscript{***}     &  -0.476\textsuperscript{***}      &     -0.604\textsuperscript{***}    \\
                               &    &  (0.005)       &  (0.004)        &   (0.004)      \\
Constructive-Analytical Percentage    &  -0.447\textsuperscript{***}    &       &     -0.410\textsuperscript{***}   &     -0.484\textsuperscript{***}     \\
                                       & (0.004)   &         &   (0.004)      &   (0.004)     \\
Questioning Percentage      &   -0.528\textsuperscript{***}   &    -0.588\textsuperscript{***}     &       &    -0.505 \textsuperscript{***}    \\
                            & (0.004)   &   (0.004)      &        &    (0.004)   \\
Critical-Evaluative Percentage     &  -0.571\textsuperscript{***}    &    -0.591\textsuperscript{***}    &  -0.430\textsuperscript{***}      &        \\
                                   &  (0.004)  &    (0.004)     &   (0.004)     &       \\
\midrule
Constant    &    50.947\textsuperscript{***}  &     61.513\textsuperscript{***}  &   43.575\textsuperscript{***}   &     54.916\textsuperscript{***}   \\
       &  (0.370)  &     (0.323)     &   (0.339)     &  (0.313)     \\

\textbf{Statistics} & & & &  \\

Adjusted R$^{2}$    & \multicolumn{1}{c}{0.500} 
                    & \multicolumn{1}{c}{0.419} 
                    & \multicolumn{1}{c}{0.362} 
                    & \multicolumn{1}{c}{0.448}\\
F-statistic      & \multicolumn{1}{c}{633.902} 
                & \multicolumn{1}{c}{891.595} 
                 & \multicolumn{1}{c}{707.333} 
                & \multicolumn{1}{c}{936.947} \\
Number of observations (N)  & \multicolumn{1}{c}{76,644} 
                     & \multicolumn{1}{c}{76,644} 
                    & \multicolumn{1}{c}{76,644} 
                    & \multicolumn{1}{c}{76,644} \\

Root MSE     & \multicolumn{1}{c}{12.215} 
        & \multicolumn{1}{c}{13.886}
        & \multicolumn{1}{c}{11.590} 
        & \multicolumn{1}{c}{12.565}\\
\bottomrule
\bottomrule
\addlinespace[1ex]

\multicolumn{3}{l}{\textsuperscript{***}$p<0.01$, 
  \textsuperscript{**}$p<0.05$, 
  \textsuperscript{*}$p<0.1$}
\end{tabular}

\end{table}

\clearpage
\begin{table}[H]
\begin{minipage}{\textwidth}
\raggedright
\scriptsize
\sisetup{table-number-alignment = center,
         table-space-text-pre ={(},
         table-space-text-post={\textsuperscript{***}},
         input-open-uncertainty={[},
         input-close-uncertainty={]},
         table-align-text-pre=false,
         table-align-text-post=false}
\begin{threeparttable}
    \caption{\textbf{Regression results by tone and reviewer identity: (1) Anonymous, (2) Disclosed.}
    Results show the estimated effects of corresponding author-related, paper-related, and review-related characteristics on the weighted proportion of each tone category (\textit{appreciative}, \textit{constructive-analytical}, \textit{questioning}, \textit{critical-evaluative}) used in peer reviews. Separate models were estimated for reviews with anonymous (1) and disclosed (2) reviewer identities. All models were estimated using ordinary least squares (OLS) regression with Huber-White robust standard errors to account for heteroscedasticity. Variance inflation factors (VIFs) indicated no multicollinearity concerns (all VIFs $<$ 6). Field of science was included as a fixed effect.}
    \label{tab:regression_results_disclosed_anon}
\begin{tabular}{@{}r 
                S[table-format=-2.3] S[table-format=-2.3] 
                S[table-format=-2.3] S[table-format=-2.3] 
                S[table-format=-2.3] S[table-format=-2.3] 
                S[table-format=-2.3] S[table-format=-2.3]@{}}

\toprule
                    & \multicolumn{2}{c}{\textbf{Appreciative}} 
                    & \multicolumn{2}{c}{\textbf{Constructive-Analytical}} 
                    & \multicolumn{2}{c}{\textbf{Questioning}} 
                    & \multicolumn{2}{c}{\textbf{Critical-Evaluative}} \\
                    & \multicolumn{1}{c}{(1)} 
                    & \multicolumn{1}{c}{(2)} 
                    & \multicolumn{1}{c}{(1)} 
                    & \multicolumn{1}{c}{(2)} 
                    & \multicolumn{1}{c}{(1)} 
                    & \multicolumn{1}{c}{(2)} 
                    & \multicolumn{1}{c}{(1)} 
                    & \multicolumn{1}{c}{(2)} \\
\midrule
\textbf{Corresponding Author Features} & & & & & & & & \\
Gender: Male                & -0.283\textsuperscript{**}   & 0.033                  & -0.340\textsuperscript{**}  & 0.234                 & -0.176                   & 0.017                 & 0.022                 & 0.418                 \\
                                & (0.107)                   & (0.224)                & (0.123)                   & (0.248)                & (0.103)                 & (0.201)                & (0.112)                & (0.213)                \\

    Race: White                   & 0.360\textsuperscript{**}   & 0.277                  & 0.159                     & -0.297                 & 0.169                   & -0.275                 & 0.017                  & -0.288                 \\
                                & (0.113)                   & (0.253)                & (0.129)                   & (0.273)                & (0.108)                 & (0.221)                & (0.118)                & (0.234)                \\

Affiliation rank: Top 100  & 0.62                     & 0.186                 & -0.644\textsuperscript{***} & -0.471                 & 0.009                   & 0.485\textsuperscript{*}   & -0.368\textsuperscript{**}  & -0.464\textsuperscript{*}   \\
                                & (0.110)                   & (0.254)                & (0.124)                   & (0.276)                & (0.106)                 & (0.227)                & (0.115)                & (0.231)                \\

Region: Western group         & 0.809\textsuperscript{***}  & 0.366                  & -1.347\textsuperscript{***} & -1.690\textsuperscript{***} & -0.260\textsuperscript{*}   & -0.236                & -1.013\textsuperscript{***} & -1.289\textsuperscript{***} \\
                                & (0.120)                   & (0.267)                & (0.138)                     & (0.289)                     & (0.116)                 & (0.233)               & (0.126)                & (0.247)                \\
                                
Academic age                  & -0.043\textsuperscript{***}& -0.033\textsuperscript{**} & -0.061\textsuperscript{***} & -0.048\textsuperscript{**} & -0.052\textsuperscript{***} & -0.031\textsuperscript{**} & -0.060\textsuperscript{***} & -0.044\textsuperscript{***} \\
                                & (0.006)                   & (0.012)                & (0.007)                   & (0.014)                & (0.006)                              & (0.011)                    & (0.006)                & (0.012)                \\

Number of prior publications       & 0.001 & -0.003 & 0.001 & -0.003 & 0.003\textsuperscript{***} & -0.001 & 0.002\textsuperscript{*} & -0.003 \\
                                & (0.001) & (0.002)  & (0.001) & (0.002)  & (0.001)   & (0.002)  & (0.001)    & (0.002)   \\
Average prior impact                    & -0.006                     & 0.002                  & -0.002                    & 0.008                  & -0.003                  & -0.005                 & -0.009\textsuperscript{*}   & 0.003                   \\
                                & (0.003)                   & (0.006)                & (0.003)                   & (0.006)                & (0.003)                 & (0.005)                & (0.004)                & (0.006)                \\
\midrule
\textbf{Paper Features}       &                            &                        &                            &                         &                          &                         &                         &                         \\

Publication year              & 0.227\textsuperscript{***} & 0.390\textsuperscript{***}  & 0.414\textsuperscript{***} & 0.575\textsuperscript{***}  & 0.122\textsuperscript{***}  & -0.027                 & -0.021                 & 0.051                  \\
                                & (0.034)                   & (0.080)                & (0.038)                   & (0.087)                & (0.032)                 & (0.071)                & (0.036)                & (0.075)                \\
Review duration
                              & -0.001                     & 0.001                  & 0.003\textsuperscript{***}  & 0.003\textsuperscript{*}    & 0.003\textsuperscript{***} & 0.004\textsuperscript{***}  & 0.010\textsuperscript{***}  & 0.008\textsuperscript{***}  \\
                                & (0.000)                   & (0.001)                & (0.000)                   & (0.001)                & (0.000)                 & (0.001)                & (0.000)                & (0.001)                \\
Team size     & -0.041\textsuperscript{**}  & 0.008   & -0.048\textsuperscript{**}  & -0.005   & 0.016    & 0.068\textsuperscript{*}    & -0.059\textsuperscript{***} & -0.029                 \\
             & (0.013)                   & (0.031)    & (0.015)                   & (0.034)    & (0.013)   & (0.028)                & (0.014)                & (0.029)                \\
             
Journal: Nature Communications & -1.679\textsuperscript{***}& 0.244                  & -6.948\textsuperscript{***} & -4.998\textsuperscript{***} & -3.350\textsuperscript{***} & -2.245\textsuperscript{***} & -3.795\textsuperscript{***} & -2.839\textsuperscript{***} \\
                                & (0.148)                   & (0.448)                & (0.162)                   & (0.483)                       & (0.143)                 & (0.393)                          & (0.156)                & (0.419)                \\
\midrule
\textbf{Review Features}      &                            &                        &                            &                         &                          &                         &                         &                         \\
Review length (words)   & -0.009\textsuperscript{***}  & -0.011\textsuperscript{***}    & 0.001\textsuperscript{***}  & -0.002\textsuperscript{***}   & 0.003\textsuperscript{***}    & 0.000    & -0.001\textsuperscript{***} & -0.002\textsuperscript{***}                 \\
                       & (0.000)                       & (0.000)                        & (0.000)                   & (0.000)                        & (0.000)                       & (0.000)   & (0.000)                & (0.000)                \\
Appreciative Percentage       &                            &                        & -0.559\textsuperscript{***} & -0.632\textsuperscript{***} & -0.474\textsuperscript{***} & -0.483\textsuperscript{***} & -0.612\textsuperscript{***} & -0.583\textsuperscript{***} \\
                              &                            &                        & (0.005)                   & (0.009)                        & (0.004)                 & (0.008)                       & (0.004)                      & (0.008)                \\
Constructive-Analytical Percentage
                              & -0.423\textsuperscript{***} & -0.518\textsuperscript{***} &                           &                         & -0.407\textsuperscript{***} & -0.419\textsuperscript{***} & -0.490\textsuperscript{***} & -0.463\textsuperscript{***} \\
                              & (0.005)                   & (0.009)                &                           &                                & (0.004)                     & (0.008)                    & (0.004)                    & (0.008)                \\
Questioning Percentage        & -0.505\textsuperscript{***}& -0.603\textsuperscript{***} & -0.573\textsuperscript{***} & -0.639\textsuperscript{***} &                          &                         & -0.510\textsuperscript{***} & -0.487\textsuperscript{***} \\
                              & (0.005)                   & (0.009)                & (0.005)                   & (0.009)                &                          &                        & (0.005)                & (0.009)                \\
Critical-Evaluative Percentage
                              & -0.549\textsuperscript{***}& -0.642\textsuperscript{***} & -0.581\textsuperscript{***} & -0.623\textsuperscript{***} & -0.430\textsuperscript{***} & -0.430\textsuperscript{***} &                          &                         \\
                              & (0.005)                   & (0.009)                & (0.005)                   & (0.010)                & (0.005)                 & (0.009)                &                         &                        \\
\midrule
Constant                      & 49.314\textsuperscript{***} & 57.844\textsuperscript{***} & 60.773\textsuperscript{***} & 65.510\textsuperscript{***} & 43.265\textsuperscript{***} & 45.135\textsuperscript{***} & 55.126\textsuperscript{***} & 53.958\textsuperscript{***} \\
                              & (0.417)                   & (0.803)                & (0.358)                   & (0.742)                & (0.378)                 & (0.777)                & (0.347)                & (0.734)                \\

\textbf{Statistics} & & & & & & & & \\
Adjusted R$^{2}$                              & \multicolumn{1}{c}{0.479} & \multicolumn{1}{c}{0.556} & \multicolumn{1}{c}{0.409} & \multicolumn{1}{c}{0.439} & \multicolumn{1}{c}{0.356} & \multicolumn{1}{c}{0.382} & \multicolumn{1}{c}{0.446} & \multicolumn{1}{c}{0.446} \\
F-statistic                                & \multicolumn{1}{c}{458.275} & \multicolumn{1}{c}{196.038} & \multicolumn{1}{c}{703.852} & \multicolumn{1}{c}{205.818} & \multicolumn{1}{c}{570.081} & \multicolumn{1}{c}{157.359} & \multicolumn{1}{c}{764.999} & \multicolumn{1}{c}{193.984} \\
Number of observations (N)                                  & \multicolumn{1}{c}{60,651} & \multicolumn{1}{c}{15,993} & \multicolumn{1}{c}{60,651} & \multicolumn{1}{c}{15,993} & \multicolumn{1}{c}{60,651} & \multicolumn{1}{c}{15,993} & \multicolumn{1}{c}{60,651} & \multicolumn{1}{c}{15,993} \\

Root MSE                                 & \multicolumn{1}{c}{11.910} & \multicolumn{1}{c}{13.121} & \multicolumn{1}{c}{13.695} & \multicolumn{1}{c}{14.502} & \multicolumn{1}{c}{11.540} & \multicolumn{1}{c}{11.742} & \multicolumn{1}{c}{12.572} & \multicolumn{1}{c}{12.494} \\
\bottomrule
\bottomrule
\end{tabular}
    \smallskip
    \scriptsize
\begin{tablenotes}[para,flushleft]
    \item[***]  $p < 0.001$
    \item[**]   $p < 0.01$,
    \item[*]    $p < 0.05$,

\end{tablenotes}
\end{threeparttable}
\end{minipage}
\end{table}

\pagebreak

\begin{table}[ht]
\centering
\caption{Distribution of key confounding variables in the full dataset and among survey respondents (corresponding authors).}
\begin{tabular}{llcc}
\hline
\textbf{Variable} & \textbf{Category} & \textbf{Full Dataset (\%)} & \textbf{Survey Respondents (\%)} \\
\hline
\multirow{2}{*}{Gender} 
  & Male   & 67.4 & 63.5 \\
  & Female & 32.6 & 36.5 \\
\hline
\multirow{2}{*}{Race} 
  & White     & 41.2 & 50.5 \\
  & Non-white & 58.8 & 49.5 \\
\hline
\multirow{2}{*}{Region} 
  & Western &  60.9 & 69.4  \\
  & Eastern & 39.1 & 30.6   \\
\hline
\multirow{2}{*}{Institution Rank} 
  & Top 100 & 31.2 & 26.5 \\
  & Other   & 68.8 & 73.5 \\
\hline
\end{tabular}
\label{tab:confounder_distribution}
\end{table}

\begin{table}[ht]
\centering
\footnotesize
\caption{OLS regression coefficients for the effects of reviewer identity, publication year, and their interaction on tone scores.}
\label{tab:interaction_regression}
\begin{tabular}{lcccc}
\toprule
\textbf{Feature} & \textbf{Appreciative} & \textbf{Constructive-Analytical} & \textbf{Questioning} & \textbf{Critical-Evaluative} \\
\midrule
Reviewer Identity
  & 1.27\textsuperscript{**} & 2.32\textsuperscript{***} & 1.15\textsuperscript{**} & --2.48\textsuperscript{***} \\
  & (0.43) & (0.45) & (0.36) & (0.42) \\
Publication Year 
  & 0.21\textsuperscript{***} & 0.62\textsuperscript{***} & --0.07 & --0.16\textsuperscript{**} \\
  & (0.05) & (0.05) & (0.04) & (0.05) \\
Interaction
 
  & 0.53\textsuperscript{***} & 0.32\textsuperscript{**} & --0.51\textsuperscript{***} & 0.01 \\
  & (0.11) & (0.11) & (0.09) & (0.11) \\
\bottomrule
Adjusted R$^{2}$   
  & 0.008 & 0.018 & 0.001 & 0.006 \\
Number of observations (N)  
  & 76,644 & 76,644 & 76,644 & 76,644 \\
\bottomrule
\multicolumn{5}{l}{\scriptsize \textsuperscript{***}$p < 0.001$, \textsuperscript{**}$p < 0.01$, \textsuperscript{*}$p < 0.05$}
\end{tabular}
\end{table}

\pagebreak
\clearpage
\subsection*{Supplementary Note 1: Sentence-Level Tone Annotation Prompt}
\label{note:prompt}

Below, we provide the prompt used with GPT-4o, along with an example peer review and the corresponding JSON output generated by the model.

\begin{tcolorbox}[enhanced,fit to height=21cm,
  colback=teal!25!black!2!white,colframe=teal!90!black,title=PROMPT, breakable, label={box:suppl_prompt_gpt}]
  
\textbf{Review Evaluation Instructions:}\\
You will assess a single peer review, evaluating it independently as feedback on your own scientific paper. Focus on the content and context. Identify the tones present in the review and assign an appropriate label to each sentence.
\begin{enumerate}[itemsep=-2pt, topsep=-1pt]
    \item \textbf{Read the Review:} Read the entire review carefully to fully understand its content and context.
    \item \textbf{Identify Contextual-Summarizing Sentences:} If a sentence provides context, summarizes, or reiterates the research without expressing any evaluative tone, label it as ``Contextual-Summarizing''.
    \item \textbf{Identify Evaluative Tones:} For all other sentences, classify them according to the relevant tone category. Handle Sarcasm Carefully: If a sentence contains sarcasm, do not classify it based on its literal meaning. Instead, determine the true intent.
    \item \textbf{Preserve Sentence Order:} The review text must be returned exactly as it appears, with each sentence tagged individually.
    \item \textbf{Determine Dominant Tone:} If a sentence exhibits multiple tones, assign the most dominant one. If two tones are equally present, list both, separated by a semicolon.
    \item \textbf{Detect and Label Signatures:}
    \begin{itemize}[itemsep=-2pt, topsep=-1pt]
        \item If the last sentence of the review contains only a name, title, or polite closing phrase (e.g., ``Sincerely, Dr. Smith''), assign the tag ``Name''.
        \item This includes sentences that contain titles such as ``Dr.'', ``Professor'', or affiliations like ``NYU, Harvard University, etc.''
    \end{itemize}
    \item \textbf{Return Output as JSON:} 
    \begin{itemize}[itemsep=-2pt, topsep=-1pt]
        \item Each sentence is a key.
        \item The assigned tone is the value.
        \item If a sentence is Contextual-Summarizing, explicitly label it as ``Contextual-Summarizing'',
        \item If a sentence does not match any tone, label it as ``None'',
        \item If the last sentence is a signature, label it as ``Name''.\\
    \end{itemize}
\end{enumerate}

\textbf{Tone Categories:}
\begin{enumerate}[itemsep=-2pt, topsep=-1pt]
    \item \textbf{Constructive-Analytical:} Provides a detailed examination of the study’s methodology, data, and results, offering targeted suggestions for improvement.
    \item \textbf{Appreciative:} This tone expresses genuine gratitude or approval, recognizing the strengths or contributions of the study.
    \item \textbf{Critical-Evaluative:} This tone highlights significant flaws, weaknesses, or limitations in the study, focusing on areas where the research does not meet standards or expectations.
    \item \textbf{Questioning:} This tone seeks further information or clarification about specific aspects of the study, indicating a need for more detail.\\
\end{enumerate}

\textbf{Output Format:}
JSON\\
\{\\
\hspace*{1em}tagged sentences:\{\\
\hspace*{2em}
``sentence 1'': ``Tone 1'',\\
\hspace*{2em}``sentence 2'': ``Tone 2'',\\
\hspace*{2em}``sentence 3'': ``Tone 3''\\
\hspace*{1em}\}\\
\}

\end{tcolorbox}

\begin{tcolorbox}[colback=blue!5!white,colframe=teal!75!black,title=Peer Review Example, breakable, label={box:suppl_peer_review_example}]
  I carefully read the submitted text. I want to emphasize its technical correctness and, above all, its content value. The authors have proposed a comprehensive method of conducting bibliometric analysis with much greater sensitivity/accuracy than previous ones. Although they themselves emphasize that the SEAC method is more time-consuming and knowledge-intensive, it is characterized by significant development potential, which the authors also write about. I hope that the SEAC method will be welcomed by the scientific community. 
\end{tcolorbox}

\begin{tcolorbox}[title=JSON Output, label={box:suppl_json_output_example_gpt}]
\begin{lstlisting}[language=json, numbers=none]
{
  "tagged_sentences": {
    "I carefully read the submitted text.": "Contextual-Summarizing",
    "I want to emphasize its technical correctness and, above all, its content value.": "Appreciative",
    "The authors have proposed a comprehensive method of conducting bibliometric analysis with much greater sensitivity/accuracy than previous ones.": "Appreciative",
    "Although they themselves emphasize that the SEAC method is more time-consuming and knowledge-intensive, it is characterized by significant development potential, which the authors also write about.": "Constructive-Analytical",
    "I hope that the SEAC method will be welcomed by the scientific community .": "Appreciative"
  }
}
\end{lstlisting}
\end{tcolorbox}

\clearpage

\pagebreak

\subsection*{Supplementary Note 2: Sample Survey Content}
\label{note:survey_content}

To validate the tone classifications generated by the GPT-4o API and ensure demographic accuracy, we conducted two survey-based validation exercises: one involving three impartial scientists, and another inviting corresponding authors to evaluate the tone of their own reviews. The primary aim was to assess the relevance and appropriateness of the assigned tone labels from both external and authorial perspectives.

Each survey began with a consent form, followed by a color-coded peer review for participants to evaluate. Participation was entirely voluntary. Respondents were provided with detailed instructions explaining the purpose and structure of the validation task.

At the end of the survey, participants were asked to report demographic information, including gender and racial or ethnic identity. This information was used to verify how well our race and gender classifiers performed.

Below, we include the full consent and instructions section of the survey. The same text was shown to both groups, with two exceptions: for impartial scientists, we omitted the introductory request confirming authorship of the review; and the prize draw section was slightly reworded, as all three impartial scientists received Amazon gift cards in appreciation for validating 100 reviews each.

\clearpage

\subsubsection*{\large{Consent and Instructions}}

Please confirm that you are the corresponding author of the following paper: \textbf{[Paper Title]}, Published in \textbf{[Journal Name]}, \textbf{[Publication year]}

\radiobtn\ \textcolor{green!50!black}{Yes}  \hspace{1cm}
\radiobtn\ \textcolor{red!70!black}{No}\\

\noindent \textbf{[If Yes, below appears. Otherwise, thanked and exits survey]}\\

\noindent Dear Dr. \textbf{[Authors' Name]},

Thank you for participating in our research study. We are investigating how factors such as authors’ gender, race, institutional affiliation, and geographical location relate to the tone of peer review reports. Your participation will help us validate our methodology and explore trends in academic publishing fairness.

In this survey, you will be shown the peer review associated with your submission titled \textbf{[Paper Title]}. The review has been split into sentences (or subsetneces), and each sentence has been color-coded to reflect one of the following tone categories based on feedback reported by GPT-4o API, a large language model:

\begin{itemize}
    \item \toneA: Expresses genuine gratitude or approval, recognizing the strengths or contributions of the study.
    \item \toneC: Provides a detailed examination of the study’s methodology, data, or findings, offering suggestions for improvement.
    \item \toneQ: Provides a detailed examination of the study’s methodology, data, or findings, offering suggestions for improvement.
    \item \toneE: Highlights flaws, weaknesses, or limitations without necessarily offering solutions.
\end{itemize}

You will be asked to evaluate each sentence individually. For each sentence:
\begin{itemize}
    \item If you agree with the assigned tone, select \textbf{“Agree.”}
    \item If you disagree, you can select the tone you believe is more appropriate from a dropdown menu and optionally provide a comment.
\end{itemize}
If you feel that more than one tone could apply to a sentence, please select the dominant tone. If you believe that two tones are equally applicable, feel free to note that in the comment section provided for that sentence.

Please note that some sentences may be labeled as \textbf{``None''} — this means no specific tone was assigned by the model. If you disagree with the ``None'' label, you can select the tone you feel best fits the sentence from the dropdown menu.

At the end of the survey, we will ask a few optional demographic questions (e.g., gender, race). Your responses will be kept strictly confidential and used only for academic research purposes. No identifying information will be published or shared beyond the research team.

We are aiming to collect responses from \textbf{500 participants}. Once this goal is reached, the survey will be closed. As a token of our appreciation, \textbf{participants who complete the survey} will be invited to enter a \textbf{random prize draw} for a chance to win one of five \$100 Amazon vouchers. \textbf{Participation is entirely optional} and will be confirmed by your response at the end of the survey. To be included in the draw, you need both \textbf{opt in} and \textbf{complete all sentence-level questions.}

Should you have any questions about either the study or your participation, you may contact Maria Sahakyan ms13502@nyu.edu or Bedoor AlShebli at bedoor@nyu.edu. For questions about your rights as a research participant, you may contact the IRB and refer to \#HRPP-2025-83, +971 2 628 4313 or IRBnyuad@nyu.edu, New York University Abu Dhabi. If you would like to have a copy of this document, please take a screenshot and keep it.

\textbf{Participation is voluntary.} By selecting an option below, you confirm that you are 18 years of age or older and agree to take part in this research study.

Your feedback is invaluable for improving our tone classification system. We sincerely appreciate your time and input.

\radiobtn\ \textcolor{green!50!black}{Yes, I agree to participate in this study.}  \hspace{1cm}
\radiobtn\ \textcolor{red!70!black}{No, I do not wish to participate.}

\clearpage

\subsubsection*{\large{Example of Review for Validation}}

The sentences in the review are color-coded to reflect the tone they convey. Refer to the table below for definitions of each tone:

\toneA \quad \toneC \quad \toneQ \quad \toneE

\vspace{1em}

\noindent\fbox{%
  \parbox{\textwidth}{%
    \textbf{Color-Coded Review:}

    \vspace{0.5em}

    \textcolor{black!50!black}{I carefully read the submitted text.}
    
    \textcolor{green!50!black}{I want to emphasize its technical correctness and, above all, its content value.}

    \textcolor{green!50!black}{The authors have proposed a comprehensive method of conducting bibliometric analysis with much greater sensitivity/accuracy than previous ones.}

    \textcolor{blue!70!black}{Although they themselves emphasize that the SEAC method is more time-consuming and knowledge-intensive, it is characterized by significant development potential, which the authors also write about.}

    \textcolor{green!50!black}{I hope that the SEAC method will be welcomed by the scientific community.}
  }%
}

\vspace{1em}
\begin{table}[H]
\centering
\renewcommand{\arraystretch}{1.5}
\begin{tabular}{|>{\raggedright\arraybackslash}m{3.8cm}|m{11cm}|}
\hline
\rowcolor{gray!10}
\textbf{Tone} & \textbf{Definition} \\
\hline

\textcolor{green!50!black}{\textbf{Appreciative}} & 
This tone expresses genuine \textbf{gratitude or approval}, recognizing the \textbf{strengths or contributions} of the study. \\
\hline

\textcolor{blue!70!black}{\textbf{Constructive-Analytical}} & 
This tone provides a \textbf{detailed examination} of the study’s methodology, data, and results, \textbf{offering targeted suggestions} for improvement. \\
\hline

\textcolor{orange!80!black}{\textbf{Questioning}} & 
This tone \textbf{seeks further information or clarification} about specific aspects of the study, indicating a need for more detail. \\
\hline

\textcolor{red!70!black}{\textbf{Critical-Evaluative}} & 
This tone \textbf{highlights} significant \textbf{flaws, weaknesses}, or \textbf{limitations} in the study, focusing on areas where the research does not meet standards or expectations. It emphasizes the identification of issues \textbf{without offering solutions}. \\
\hline

\end{tabular}
\end{table}

\begin{tabular}{|m{6cm}|>{\centering\arraybackslash}m{3.2cm}|>{\centering\arraybackslash}m{2.4cm}|m{1.9cm}|}

\hline
\textbf{Sentence} &
\textbf{Do you agree?} \newline \textcolor{green!50!black}{Agree} \hspace{1em} \textcolor{red}{Disagree} &
\textbf{If not, select the correct tone} &
\textbf{Optional comment} \\
\hline

\textit{I carefully read the submitted text.} \mbox{\textbf{None}} &
\radiobtn \hspace{2em} \radiobtn &
\makecell{\fbox{\textcolor{gray}{Select one}}} &  \\

\hline
\textit{I want to emphasize its technical correctness and, above all, its content value.} \mbox{\toneA} &
\radiobtn \hspace{2em} \radiobtn &
\makecell{\fbox{\textcolor{gray}{Select one}}} & \\

\hline
\textit{The authors have proposed a comprehensive method of conducting bibliometric analysis with much greater sensitivity/accuracy than previous ones.} \toneA &
\radiobtn \hspace{2em} \radiobtn &
\makecell{\fbox{\textcolor{gray}{Select one}}} & \\
\hline

\textit{Although they themselves emphasize that the SEAC method is more time-consuming and knowledge-intensive, it is characterized by significant development potential, which the authors also write about.} \mbox{\toneC} &
\radiobtn \hspace{2em} \radiobtn &
\makecell{\fbox{\textcolor{gray}{Select one}}} & \\
\hline
\textit{I hope that the SEAC method will be welcomed by the scientific community.} \mbox{\toneA} &
\radiobtn \hspace{2em} \radiobtn &
\makecell{\fbox{\textcolor{gray}{Select one}}} & \\
\hline
\end{tabular}

\pagebreak

\subsection*{Supplementary Note 3: Illustrative Comparison of Tone Weighting Methods by Sentence Length}
\label{note:sentence_word_length}

To demonstrate the importance of incorporating sentence length into tone scoring, we present two peer review examples. Both contain 10 sentences, of which 2 are labeled as Appreciative, yet the nature of appreciation differs significantly between the two.

The first review includes brief and generic appreciative statements (e.g., “This article is well written”), while the second review offers more substantive engagement with the manuscript’s contributions. If both were scored equally by sentence frequency alone, these differences would be obscured.

To address this, we applied the hybrid tone weighting method described in the Methods section. This approach modestly adjusts tone scores to reflect the elaboration of tone, without overpowering the overall distribution. The example below includes color-coded sentences, with tones assigned as follows:\\

\toneA \quad \toneC \quad \toneQ \quad \toneE
\vspace{0.5em}\\

\noindent
\textbf{Example 1: Short Review with Brief Appreciative Sentences}\\

\noindent\fbox{%
  \parbox{\textwidth}{%

    \textcolor{green!50!black}{This article is well written.} 
    \textcolor{green!50!black}{Methodology seems sound.}
    \textcolor{red!70!black}{Conclusions were pretty predictable.}
     \textcolor{orange!80!black}{I have three questions.}
     \textcolor{orange!80!black}{1- Could the authors please elaborate on the fact that vaccination was not available on sit everywhere ?}
     \textcolor{red!70!black}{It seems odd that it is not the case in every setting.}
    \textcolor{orange!80!black}{could you please describe what health care workers have to do to be vaccinated when flu vaccine is not available on site ?}
    \textcolor{orange!80!black}{2- Do they authors have the 2020/2021 national vaccination coverage in physicians and nurses as a comparator ?}
    \textcolor{orange!80!black}{3- Could you precise others categories than administrative personnel in the non physicians or nurses people ?}
    \textcolor{orange!80!black}{Did you survey other personnels in contact with patients such as cargivers known for their low vaccine coverage (even lower than nurses’ one) ?}
  }%
}

\vspace{0.5em}
\noindent
\textbf{Example 2: Longer Review with Substantive Appreciative Sentences}\\


\noindent\fbox{%
  \parbox{\textwidth}{%
    \textcolor{black!70!black}{This manuscript proposed a background correction method to compensate for the variations in surface morphology or light power distribution at the sample.} 
    \textcolor{green!50!black}{The proposed background correction method enables the estimation of optical characteristics of illumination at the sample, which is a major advance in hyperspectral endoscopy systems.} 
    \textcolor{black!70!black}{The authors experimentally demonstrated the feasibility of the proposed background correction method using hyperspectral imaging data acquired via a hyperspectral endoscopy system from different samples.}
    \textcolor{green!50!black}{The proposed method and experiment results are useful in further exploitation of hyperspectral imaging endoscopy systems in practical clinical applications, the manuscript can be accepted for publication.} 
    \textcolor{orange!80!black}{A few questions that should be addressed:'}

    \textcolor{orange!80!black}{1. Where is the multiplicative factor alpha in eq. (1) and constant in eq. (2) ?}
    
    \textcolor{orange!80!black}{2. In the result section p5, line 10. Did you multiply the normalized spectrum of the light source with the intensity ratio between the measured spectral profiles of the light source and the sample or with the intensity ratio between the normalized spectral profiles of light source and sample?}
    
    \textcolor{orange!80!black}{3. The wavelength 800 nm was selected as the band with low absorption to verify the feasibility of the proposed method. the question is, if the absorption of a certain tissue is not that small at 800nm, does the method still work? if not, how to solve this problem?}
    
    \textcolor{red!70!black}{4. In figure 5a, there is no much difference in darkness between rgb images of gt and sb? in contrast, gt and sb looks similar.}
        
    \textcolor{red!70!black}{5. In figure 5a, six colored rectangular boxes are obscure.}

  }%
}\\

This hybrid weighting ensures that both short, high-impact tone cues and longer, detailed feedback are fairly represented, offering a more accurate and nuanced measure of the tone conveyed in peer review texts.

\pagebreak
\subsection*{Supplementary Note 4: Reviewer Tone Over Time and the Role of Identity Disclosure}
\label{note:tone_temporal}

During our analysis, we observed that publication year was a consistent and significant predictor of tone scores in both the full models and the regressions stratified by reviewer identity. While this does not establish a clear trend, it signals that reviewer tone may vary across time. At the same time, we found systematic differences in tone between anonymous and disclosed reviewers, which are likely shaped by differences in visibility and accountability. Based on these observations, we hypothesized that the way tone evolves over time may itself depend on whether or not reviewers disclose their identities. In particular, we expected that disclosed reviewers may increasingly adopt more supportive and collegial language in response to evolving community norms and reputational incentives. To test this hypothesis, we estimated a series of regression models including interaction terms between publication year and reviewer identity, allowing us to assess whether tone trajectories over time differ between anonymous and disclosed reviews.

To assess whether tone trajectories over time differ between anonymous and disclosed peer reviews, we estimated a series of linear regression models separately for each tone category. Each model included terms for publication year (scaled from 1 to 6), reviewer identity disclosure (coded as 0 for anonymous and 1 for disclosed), and their interaction, to capture whether the rate of change in tone over time differs by reviewer identity disclosure status. The regression model took the following form:
\begin{equation}
\text{Tone}_i = \beta_0 + \beta_1 \cdot \text{Reviewer identity}_i + \beta_2 \cdot \text{Year}_i + \beta_3 \cdot (\text{Reviewer identity}_i \times \text{Year}_i) + \varepsilon_i
\end{equation}

Here, $\text{Tone}_i$ is the weighted tone score for observation $i$, $\text{Reviewer identity}_i$ indicates the disclosure status of the reviewer's identity (1 = disclosed, 0 = anonymous), and $\text{Year}_i$ is the scaled publication year. The coefficient $\beta_3$ on the interaction term captures whether the trajectory of tone over time differs between anonymous and disclosed reviewers.

Across all models, the interaction term captured whether the temporal trend in tone varied significantly between the two reviewer groups. The results revealed tone-specific interaction effects. 

The regression results indicate that the disclosure of the reviewer's identity is associated with distinct temporal changes in tone. Specifically, we observed a significant positive interaction between publication year and reviewer identity for both appreciative and constructive-analytical tones. This suggests that disclosed reviewers are not only more likely to use appreciative and constructive language, but that these tone dimensions have become increasingly prevalent over time in disclosed reviews. In contrast, anonymous reviews showed a relatively flatter or more modest increase. While we cannot establish causality, these patterns, observed consistently in both the stratified models presented in the main text and the interaction models reported here, suggest that identity disclosure is associated with an increasing use of more supportive and constructive tone over time. These trends can also be observed in Supplementary Figure~\ref{fig:suppl_tone_over_time}.

For questioning tone, the interaction effect was negative and statistically significant, indicating that disclosed reviewers have become less likely to include probing or interrogative comments over time, more so than their anonymous counterparts. This could reflect a shift away from critical interrogation in favor of more collaborative engagement when reviewer identities are known. Finally, no significant interaction was found for critical-evaluative tone, suggesting that the level of critique remains relatively stable over time and does not vary significantly by reviewer identity. 

Taken together, these findings suggest that the increasing use of appreciative and constructive language over time is largely driven by disclosed reviewers. In contrast, the questioning tone shows a notable decline over time specifically in disclosed reviews, while the critical tone remains relatively stable across both reviewer groups.

\end{document}